\documentclass[times,twocolumn,preprint,final,authoryear]{elsarticle}

\usepackage{ycviu}
\usepackage{framed,multirow}

\usepackage{amsmath,amssymb} %
\usepackage{latexsym}
\usepackage[linesnumbered,vlined,ruled]{algorithm2e}

\usepackage{url}
\definecolor{newcolor}{rgb}{.8,.349,.1}
\usepackage[dvipsnames]{xcolor}

\newcommand{\ie}{\textit{i}.\textit{e}.}
\newcommand{\eg}{\textit{e}.\textit{g}.}
\usepackage{pifont}

\journal{Preprint}
\begin{document}

\ifpreprint
  \setcounter{page}{1}
\else
  \setcounter{page}{1}
\fi

\begin{frontmatter}

\title{Lightning Fast Video Anomaly Detection via Multi-Scale Adversarial Distillation}

\author[1]{Florinel-Alin \snm{Croitoru}}
\author[1,2]{Nicolae-C\u{a}t\u{a}lin \snm{Ristea}} 
\author[1]{Dana \snm{D\u{a}sc\u{a}lescu}}
\author[1,3]{Radu Tudor \snm{Ionescu}\corref{cor1}}
\cortext[cor1]{Corresponding author at: Department of Computer Science, University of Bucharest, 14 Academiei Street, Bucharest 010014, Romania}
\ead{raducu.ionescu@gmail.com}
\author[4, 5]{Fahad Shahbaz \snm{Khan}}
\author[6]{Mubarak \snm{Shah}}

\address[1]{Department of Computer Science, University of Bucharest, 14 Academiei Street, Bucharest 010014, Romania}
\address[2]{National University for Science and Technology Politehnica Bucharest, 313 Splaiul Independenței Street, Bucharest, Romania}
\address[3]{SecurifAI, 21D Mircea Voda, Bucharest 030662, Romania}
\address[4]{MBZ University of Artificial Intelligence, Masdar City, Abu Dhabi, UAE}
\address[5]{Link\"{o}ping University, 581 83 Link\"{o}ping, Sweden}
\address[6]{Center for Research in Computer Vision (CRCV), University of Central Florida, Orlando 32816, FL, US}


\begin{abstract}
We propose a very fast frame-level model for anomaly detection in video, which learns to detect anomalies by distilling knowledge from multiple highly accurate object-level teacher models. To improve the fidelity of our student, we distill the low-resolution anomaly maps of the teachers by jointly applying standard and adversarial distillation, introducing an adversarial discriminator for each teacher to distinguish between target and generated anomaly maps. We conduct experiments on three benchmarks (Avenue, ShanghaiTech, UCSD Ped2), showing that our method is over 7 times faster than the fastest competing method, and between 28 and 62 times faster than object-centric models, while obtaining comparable results to recent methods. Our evaluation also indicates that our model achieves the best trade-off between speed and accuracy, due to its previously unheard-of speed of 1480 FPS. In addition, we carry out a comprehensive ablation study to justify our architectural design choices. Our code is freely available at: \url{https://github.com/ristea/fast-aed}.
\end{abstract}

\begin{keyword}
\MSC[2008] 68T01\sep 68T45\sep 68U10\sep 62M45
\KWD abnormal event detection\sep anomaly detection\sep knowledge distillation\sep neural networks \sep transformers
\end{keyword}

\end{frontmatter}

\section{Introduction}

\begin{figure}[!t]
\begin{center}
\centerline{\includegraphics[width=1.0\linewidth]{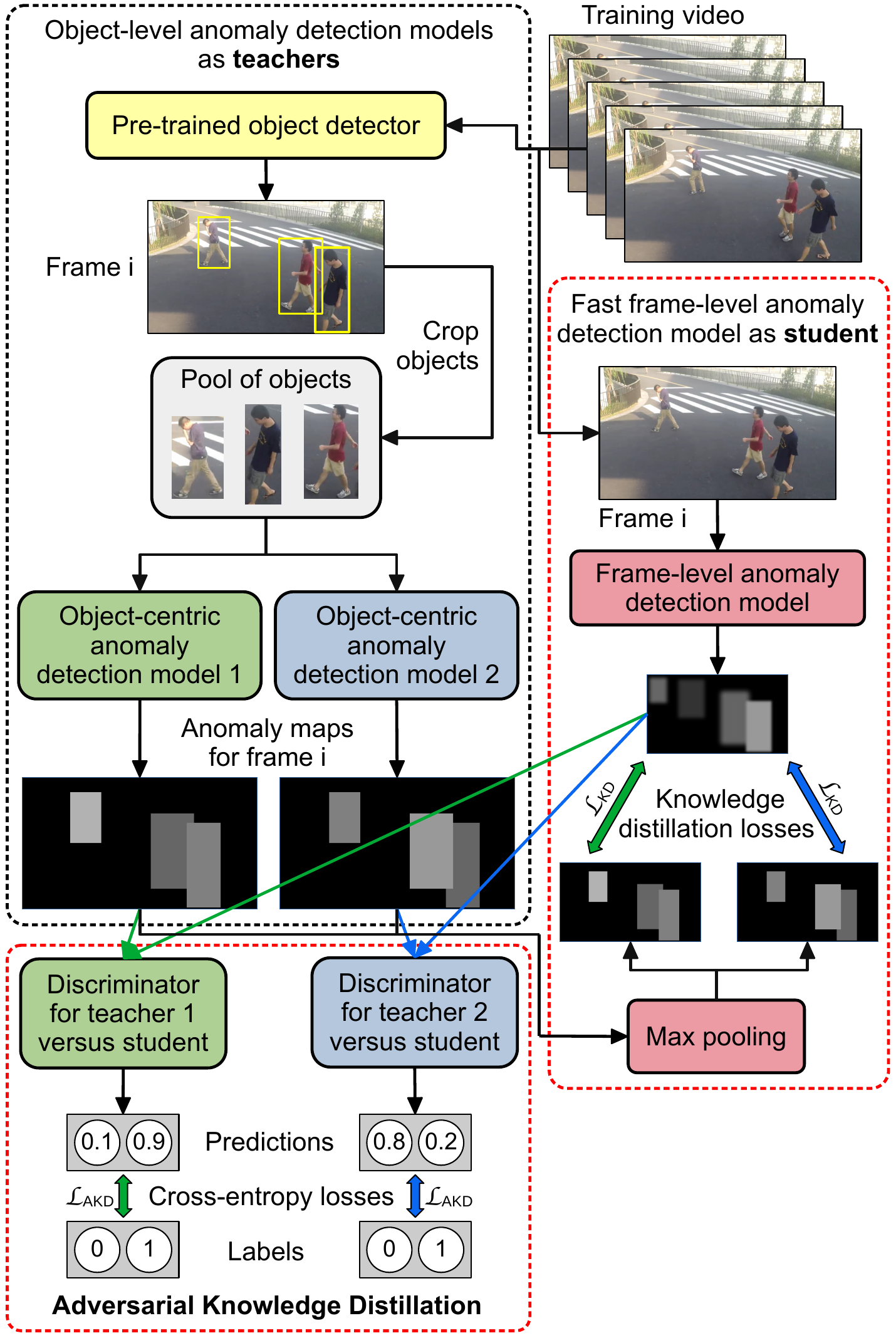}}
\caption{Our pipeline comprises an efficient student model that learns to distill knowledge from two object-centric teachers via a combination of direct and adversarial losses. The pipeline can be trivially extended to any number of teachers. The knowledge distillation loss $\mathcal{L}_{\mbox{\scriptsize{KD}}}$ is defined in Eq.~\eqref{pixel_distillation}, while the adversarial knowledge distillation loss $\mathcal{L}_{\mbox{\scriptsize{AKD}}}$ is defined in Eq.~\eqref{adv_distillation}. For simplicity, we do not represent the multi-frame input and multi-resolution output of our student. Best viewed in color.}
\label{fig_pipeline}
\end{center}
\end{figure}

Video anomaly detection is a difficult task, being actively studied in recent years \citep{Acsintoae-CVPR-2022,Doshi-CVPRW-2020a,Georgescu-CVPR-2021,Georgescu-TPAMI-2021,Gong-ICCV-2019,Ionescu-WACV-2019,Ionescu-ICCV-2017,Li-ECCV-2022,Liu-ICCV-2021,Lu-ECCV-2020,Nguyen-ICCV-2019,Pang-CVPR-2020,Ramachandra-PAMI-2020,Ristea-CVPR-2022,Smeureanu-ICIAP-2017,Sun-ACMMM-2020,Tang-PRL-2020,Vu-AAAI-2019,Wang-ACMMM-2020,Yu-ACMMM-2020,Yu-CVPR-2022,Zaheer-CVPR-2022} due to its increasing importance. The difficulty of the task stems from the fact that abnormal events depend on the context and occur rarely. To illustrate the scarcity and reliance on context of abnormal events, we consider the truck deliberately driven into the Berlin Christmas market\footnote{\url{https://en.wikipedia.org/wiki/2016_Berlin_truck_attack}} in 2016 as a relevant example. A truck driven on the road is labeled as a normal event, but, as soon as the truck intrudes into a pedestrian area, \eg~a market, the event becomes abnormal. This example confirms that labeling an event as normal or abnormal depends on the context. Moreover, according to Wikipedia\footnote{\url{https://en.wikipedia.org/wiki/Vehicle-ramming_attack}}, there are about 170 vehicle ramming attacks registered since 1953 to date, but many of them are too old to be caught on video. Hence, our example also confirms that such events rarely occur.

Due to the rarity and reliance on context of abnormal events, it is nearly impossible to collect a sufficiently representative set of such events, eliminating the possibility of applying traditional approaches based on supervised learning to perform video anomaly detection. Without the supervised option at hand, researchers turned their attention to alternative solutions, the most prominent alternative being based on outlier detection \citep{Cong-CVPR-2011,Dutta-AAAI-2015, Hasan-CVPR-2016,Ionescu-WACV-2019,Kim-CVPR-2009,Lee-TIP-2019,Liu-CVPR-2018,Lu-ICCV-2013,Luo-ICCV-2017,Park-CVPR-2020,Ramachandra-WACV-2020a,Ramachandra-WACV-2020b,Ramachandra-PAMI-2020,Sabokrou-IP-2017,Wu-TNNLS-2019,Zhong-CVPR-2019}. Methods based on outlier detection learn a normality model from training videos containing only normal events, labeling test samples that deviate from the model as abnormal. Our approach falls within the same line of research, being based on distilling knowledge from object-centric outlier detection systems \citep{Georgescu-CVPR-2021,Georgescu-TPAMI-2021}, which reported state-of-the-art performance levels in anomaly detection. We hereby emphasize that top scoring abnormal event detection systems such as \citep{Barbalau-CVIU-2023,Georgescu-CVPR-2021,Georgescu-TPAMI-2021,Ionescu-CVPR-2019,Liu-ICCV-2021} operate at the object level, relying on a pre-trained object detector which limits the real-time processing to only a single video stream per GPU. Although it might seem sufficient to process one video stream at about 20-30 FPS on one GPU, the cost of computational resources to process tens or hundreds of video streams (as expected in real-world surveillance scenarios) quickly becomes very high, considering that the price range of one GPU is around \$1000-4000. The high cost of the computational resources adds to the difficulty of solving the video anomaly detection task in the real-world.

\begin{figure}[!t]
\begin{center}
\centerline{\includegraphics[width=1.0\linewidth]{fig_tradeoff_av.pdf}}
\caption{The trade-off between performance (micro AUC) and speed (FPS) for our student versus multiple state-of-the-art methods \citep{Georgescu-CVPR-2021,Georgescu-TPAMI-2021,Gong-ICCV-2019,Ionescu-ICCV-2017,Liu-CVPR-2018,Liu-ICCV-2021,Park-CVPR-2020,Park-WACV-2022,Ristea-CVPR-2022,Wang-ECCV-2022} (with available code), on the Avenue data set. The running times of all methods are measured on a machine with an Nvidia GeForce GTX 3090 GPU with 24 GB of VRAM. Best viewed in color.}
\label{fig_tradeoff_avenue}
\end{center}
\end{figure} 

To address the aforementioned efficiency problem, we propose a fast anomaly detection model based on a novel convolutional transformer that replaces fully connected layers inside the transformer with pointwise convolutional layers. Our model operates at the frame level, eliminating the need for a pre-trained object detector. At the same time, we aim to reach an accuracy level that is as close as possible to that of object-centric models. To this end, we train our model in a teacher-student setup, learning to detect anomalies by distilling knowledge from multiple highly accurate object-centric teachers. Aside from the conventional distillation process based on the mean squared error between the low-resolution anomaly maps of the student and each teacher, we introduce an adversarial discriminator to distinguish between target (teacher) and reproduced (student) anomaly maps, as shown in Figure \ref{fig_pipeline}. To the best of our knowledge, we are the first to propose an adversarial knowledge distillation framework for anomaly detection in video. 

We conduct experiments on three benchmark data sets: Avenue \citep{Lu-ICCV-2013}, ShanghaiTech \citep{Luo-ICCV-2017} and UCSD Ped2 \citep{Mahadevan-CVPR-2010}. Our empirical evaluation shows that our method is 28 to 62 times faster compared with object-centric models (1480 FPS versus 24-52 FPS), while obtaining performance levels that are comparable to many recent methods \citep{Astrid-BMVC-2021, Astrid-ICCVW-2021, Barbalau-CVIU-2023, Doshi-CVPRW-2020a,  Ionescu-CVPR-2019, Ionescu-WACV-2019, Liu-ICCV-2021, Park-CVPR-2020, Ramachandra-WACV-2020b, Sun-ACMMM-2020, Tang-PRL-2020, Yu-TNNLS-2021} and less than $5\%$ below the object-centric teachers \citep{Georgescu-CVPR-2021,Georgescu-TPAMI-2021}. As shown in Figure \ref{fig_tradeoff_avenue}, our model achieves an excellent trade-off between speed and accuracy. In addition, we carry out a comprehensive ablation study to justify our architectural design choices. 

In summary, our contribution is fourfold:
\begin{itemize}
    \item We propose a novel teacher-student framework for anomaly detection in video, learning to detect anomalies by distilling from multiple highly accurate object-level teachers into a highly efficient student, at multiple scales.
    \item We propose adversarial knowledge distillation in the context of anomaly detection, introducing an adversarial discriminator for each teacher to distinguish between original (teacher) and reproduced (student) outputs.    
    \item We carry out comparative and ablation experiments to assess the performance and speed gains of our design.
    \item To further increase the processing speed, we replace the fully connected layers inside the transformer blocks forming the student with pointwise convolutional layers.
\end{itemize}

\section{Related work}

Most frameworks formulate anomaly detection as a one-class learning problem, where only normal data is available at training time, while both normal and abnormal examples are present at test time \citep{Pang-CSUR-2021,Ramachandra-PAMI-2020}. The anomaly detection approaches are typically categorized into dictionary learning methods \citep{Cheng-CVPR-2015,Cong-CVPR-2011,Dutta-AAAI-2015,Lu-ICCV-2013}, probabilistic models \citep{Feng-NC-2017,Hinami-ICCV-2017,Saleh-CVPR-2013,Wu-CVPR-2010}, distance-based approaches \citep{Ionescu-CVPR-2019,Ionescu-WACV-2019,Ramachandra-WACV-2020a,Ramachandra-WACV-2020b,Sabokrou-IP-2017,Sabokrou-CVIU-2018,Saligrama-CVPR-2012,Smeureanu-ICIAP-2017}, reconstruction-based methods \citep{Fei-TMM-2020,Gong-ICCV-2019,Hasan-CVPR-2016,Liu-CVPR-2018,Nguyen-ICCV-2019,Park-CVPR-2020,Ristea-CVPR-2022} and change detection frameworks \citep{Giorno-ECCV-2016,Ionescu-ICCV-2017,Liu-BMVC-2018,Pang-CVPR-2020}. 
Anomaly detection methods can also be divided into frame-level \citep{Ionescu-WACV-2019,Liu-CVPR-2018,Lu-ICCV-2013,Mahadevan-CVPR-2010,Mehran-CVPR-2009,Sabokrou-IP-2017,Saligrama-CVPR-2012,Yu-TNNLS-2021} and object-level \citep{Barbalau-CVIU-2023,Doshi-CVPRW-2020a,Georgescu-CVPR-2021,Georgescu-TPAMI-2021,Ionescu-CVPR-2019,Liu-ICCV-2021,Wang-ECCV-2022,Yu-ACMMM-2020} frameworks, according to the level at which the anomaly detection algorithm is applied. These two categories are detailed below.

\noindent
\textbf{Frame-level methods.} 
Frame-level methods detect anomalies by taking entire video frames as input. \citet{Yu-TNNLS-2021} proposed a frame-level framework that employs the adversarial learning of past and future events to detect anomalies. 
\citet{Liu-CVPR-2018} proposed a simple yet effective algorithm, which learns to reconstruct the next frame of a short video sequence. 
Considering that frame-level methods commonly rely on producing poor reconstructions for anomalies, generalizing to out-of-distribution samples, \eg~anomalies, is not desired. To reduce the generalization capability, researchers employed various techniques, such as adding memory modules \citep{Gong-ICCV-2019} or masked convolutional blocks \citep{Madan-ARXIV-2022,Ristea-CVPR-2022}. Although integrating additional modules into the network leads to accuracy gains, the procedure often comes with a significant efficiency penalty. 

Focusing primarily on efficiency, we propose a novel adversarial distillation framework that teaches our efficient frame-level student to replicate two heavy object-level teachers \citep{Georgescu-CVPR-2021,Georgescu-TPAMI-2021}. By distilling powerful teachers, we can afford to employ a lightweight frame-level student, gaining between 28 and 62 times the speed of the teachers with just a small performance drop (less than $5\%$).

\noindent
\textbf{Object-level methods.}
Lately, several powerful algorithms have been developed for anomaly detection by estimating anomaly scores at the object level \citep{Barbalau-CVIU-2023,Ionescu-CVPR-2019, Georgescu-TPAMI-2021, Georgescu-CVPR-2021,Liu-ICCV-2021,Yu-ACMMM-2020,Wang-ECCV-2022}, rather than at the frame level. Since the abnormal examples can be described as unusual objects in a regular scene, the prior information from an object detector enables anomaly detection models to focus only on objects, boosting the performance by a significant margin. 
One of the first object-level models \citep{Ionescu-CVPR-2019} is based on applying a k-means clustering in the joint latent space of motion and appearance auto-encoders. Later, \citet{Georgescu-TPAMI-2021} extended the framework by introducing pseudo-anomalies during training and attaching classifiers to discriminate between normal and pseudo-abnormal samples.
\citet{Georgescu-CVPR-2021} employed self-supervised multi-task learning at the object-level to detect anomalies in video. 
The approach is further developed in a study \citep{Barbalau-CVIU-2023} that adds new proxy tasks and a transformer-based architecture. \citet{Wang-ECCV-2022} studied alternative proxy tasks based on solving spatial and temporal jigsaw puzzles. A more complex approach is proposed by \citet{Liu-ICCV-2021}, which conditions the image prediction on optical flow reconstruction using a conditional Variational Auto-Encoder (VAE).


Object-level methods rely on running an object detector, which limits the processing time to the speed of the detector. Thus, efficiency is significantly impacted, as the speed of the detector is often much lower than that of the anomaly network \citep{Ionescu-CVPR-2019,Georgescu-CVPR-2021}. To address this issue, we propose a frame-level framework that is trained to reproduce the output of two strong object-level teachers \citep{Georgescu-TPAMI-2021, Georgescu-CVPR-2021}, eliminating the object detector from the equation during inference.

\noindent
\textbf{Methods addressing efficiency.}
While video anomaly detection has been extensively studied in recent years, the top scoring methods have certain speed limitations, preventing the deployment of state-of-the-art approaches in real-world scenarios, \eg~in video surveillance of entire cities. Even so, there is limited research on efficient anomaly detection frameworks \citep{Doshi-CVPR-2021, Fang-TM-2020, Li-CVIU-2021, Park-WACV-2022}, and an acceptable trade-off between speed and performance has not been achieved yet. \citet{Doshi-CVPR-2021} proposed a video anomaly detection system that is capable of running on roadside cameras, 
while \citet{Li-CVIU-2021} proposed a neural architecture that learns normal behavior without supervision. 

In our work, we design a lightweight hybrid convolutional-transformer student network that is trained to mimic the output of highly complex and accurate teachers, in an adversarial fashion. The limited size of our student leads to processing more than 1480 frames per second (FPS), while achieving competitive accuracy. Compared with competing methods, we achieve a significantly better speed versus performance trade-off.

\noindent
\textbf{Transformers in anomaly detection.}
Transformers have been widely adopted in computer vision, due to the outstanding results across a broad variety of tasks, ranging from object recognition \citep{Dosovitskiy-ICLR-2020,Wu-ICCV-2021} and object detection \citep{Carion-ECCV-2020,Zhu-ICLR-2020} to image generation \citep{Ristea-A-2021,Xu-AAAI-2022}. Distinct from approaches using only transformer-based attention \citep{Carion-ECCV-2020,Chen-arXiv-2021,Dosovitskiy-ICLR-2020,Parmar-ICML-2018,Ristea-A-2021,Vaswani-NIPS-2017,Wu-ICCV-2021,Zhu-ICLR-2020}, we propose a hybrid model, composed of both convolutional and self-attention blocks, to reduce the usually high processing time of transformers. More precisely, we improve the efficiency by downsampling the input and by replacing fully connected layers with pointwise convolutional layers. 

To the best of our knowledge, there are only a handful of works applying transformers to video anomaly detection \citep{Barbalau-CVIU-2023,Jin-TGRS-2022, Madan-ARXIV-2022, Chen-NN-2022}. Unlike these related methods, which barely discuss the efficiency problem, our work is the first to significantly improve the accuracy versus efficiency trade-off of transformers for anomaly detection. 


\begin{figure*}[!t]
\begin{center}
\centerline{\includegraphics[width=1.0\linewidth]{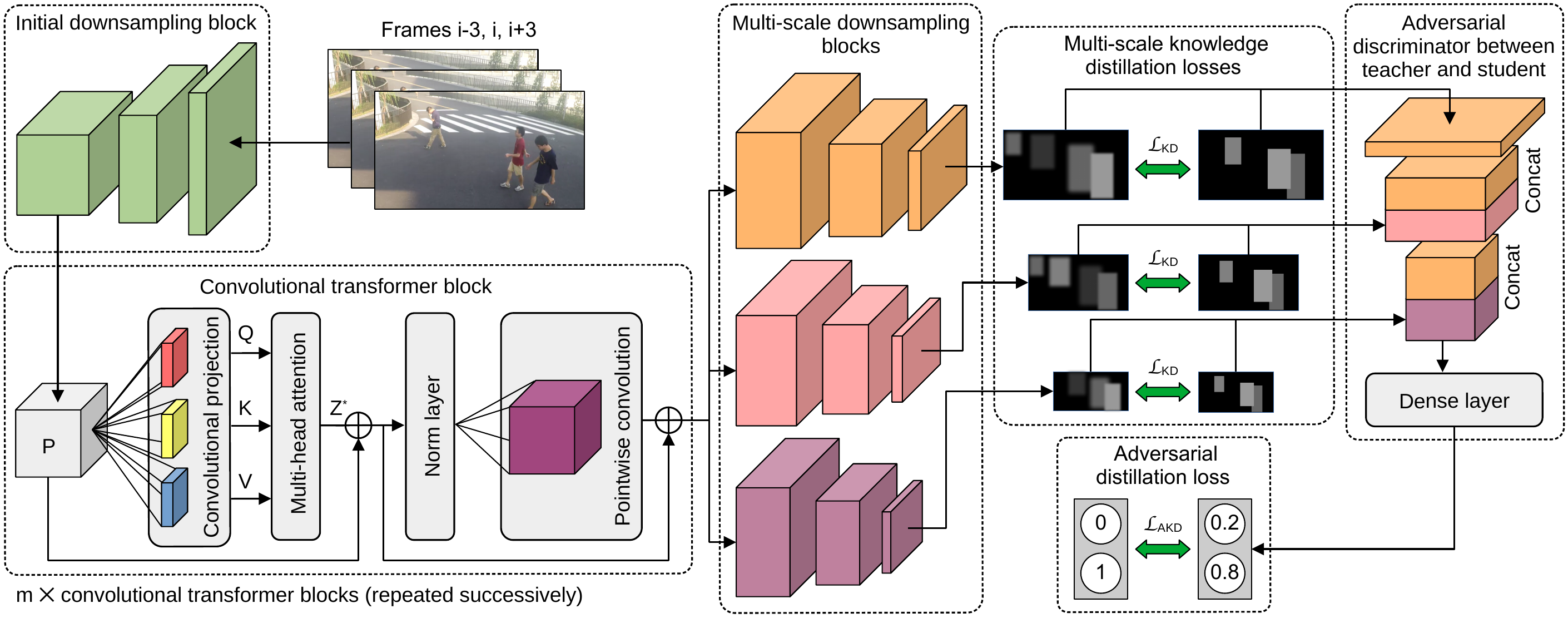}}
\caption{Architecture of our student based on a convolutional transformer and multiple downsampling blocks. For simplicity, we represent the adversarial discriminator and anomaly maps for only one teacher. Best viewed in color.}
\label{fig_arch}
\end{center}
\end{figure*}

\noindent
\textbf{Knowledge distillation in anomaly detection.}
As a strategy for compressing one or multiple highly complex models, knowledge distillation \citep{Ba-NIPS-2014,Hinton-DLRL-2015} aims to efficiently train a lightweight neural network, called \emph{student}, under the supervision of one or more deeper neural networks, called \emph{teachers}, achieving competitive performance with fewer computational resources. Knowledge distillation was recently adopted for various vision tasks \citep{Gou-IJCV-2021}, \eg~object recognition \citep{Chung-ICML-2020,Park-CVPR-2019,Yu-CVPR-2019}, face recognition \citep{Feng-ICIP-2020,Georgescu-MVA-2022} and anomaly detection in industrial images \citep{Bergmann-CVPR-2020,Cheng-PRCV-2021,Deng-CVPR-2022,Salehi-CVPR-2021}. 
Comparatively lower attention has been devoted to applying knowledge distillation to video anomaly detection \citep{Georgescu-CVPR-2021,Wang-MLSP-2021}. \citet{Wang-MLSP-2021} employed a self-taught system based on a teacher-student model that does not require human-annotated data. Similarly, \citet{Georgescu-CVPR-2021} employed knowledge distillation and integrated it with other self-supervised tasks. 
The methods employing knowledge distillation for image or video anomaly detection leverage the representation discrepancy of anomalies between the teacher and the student models, which requires running both teacher and student models during inference. 

Unlike the aforementioned methods, we perform knowledge distillation from teachers designed for video anomaly detection, specifically aiming to improve the computational time, running only the student model at test time. 
Moreover, to the best of our knowledge, we are the first to introduce adversarial knowledge distillation in the context of anomaly detection.

\section{Method}

\noindent
\textbf{Overview.}
We propose to train a fast and competitive frame-level model for video anomaly detection via multi-scale distillation. The training process is divided into two phases. In the first training phase, we attach a decoder to our network and train it as an auto-encoder to reconstruct normal training data, in a similar manner to previous studies \citep{Lee-TIP-2019,Liu-CVPR-2018}. Then, we diverge from the commonly-used auto-encoder model by discarding the decoder, attaching three anomaly detection heads to the model and training it via knowledge distillation. We employ two strategies to distill the knowledge from two object-level anomaly detection teacher models \citep{Georgescu-CVPR-2021,Georgescu-TPAMI-2021} into our student, as illustrated in Figure~\ref{fig_pipeline}. The first strategy is based on minimizing the mean squared error between the target anomaly maps provided by the teachers and the anomaly maps predicted by the student. The second strategy is based on adversarial training, relying on a neural discriminator that is optimized to distinguish between student and teacher anomaly maps. At the same time, our student is trained with an adversarial loss, trying to fool the discriminator. In this training setting, the student becomes a conditional generator. 

An anomaly map is a matrix containing anomaly scores, which indicates the location of anomalies in a video frame. The teachers detect anomalies at the object level. To construct the teacher anomaly maps, we start from a map initialized with zeros, which is the same size as the corresponding video frame. For each detected object, we fill the corresponding bounding box in the anomaly map with the anomaly score predicted by a teacher. Following \citep{Georgescu-CVPR-2021,Georgescu-TPAMI-2021}, we keep the maximum anomaly score when two bounding boxes overlap. Finally, a Gaussian blur filter is applied to smooth the anomaly maps. These anomaly maps are downsampled at multiple scales (using bilinear interpolation) and used as targets for the student model.
Our student architecture takes video frames as input and directly predicts the anomaly maps. This allows the student to bypass the costly object detection step. We next present the architecture and the training procedure of our student network.

\noindent
\textbf{Student architecture.}
As depicted in Figure~\ref{fig_arch}, the architecture of our student network starts with a downsampling block which reduces the spatial dimensions of the input frames, thus speeding-up the subsequent processing. The second part of the architecture is formed of $m$ adapted convolution vision transformer (CvT) \citep{Wu-ICCV-2021} blocks. The adaptation consists of replacing all dense layers with pointwise convolutional layers to further speed up the computation. The last part incorporates three multi-scale downsampling blocks generating anomaly maps at various resolutions.

Our initial downsampling block includes five convolutional layers, each followed by batch normalization and Rectified Linear Units (ReLU) \citep{Nair-ICML-2010}. 
Let $P \in \mathbb{R}^{h\times w\times c}$ denote the output tensor of the downsampling block, where $h$, $w$ and $c$ represent the height, width and number of channels. $P$ is further processed by a convolutional transformer. The transformer encloses $m$ consecutive transformer blocks and interprets $P$ as a set of overlapping visual tokens. The transformer blocks follow the implementation of \citet{Wu-ICCV-2021}, where, prior to the self-attention mechanism, the queries $Q$, the keys $K$ and the values $V$ are computed from $P$ via convolutional projection. 
The next stage in the transformer block is the self-attention mechanism, which computes a new value vector for each token as a weighted sum of all value vectors in the set, where the weights are the attention scores. Formally, the operation is defined as follows:
\begin{equation}
 \label{eq_self_attention}
    \begin{split}
        Z = \mbox{\emph{softmax}}\left(\frac{
        Q\cdot K^\top}{\sqrt{d_q}}\right)\cdot V,
    \end{split}
\end{equation}
where $d_q$ is the size of a query token, $K^\top$ is the transpose of $K$, and $Z$ is a matrix containing the new value vectors. We reshape $Z$ into a tensor of $h\times w\times d$ components.
Our network comprises a multi-head attention module comprising $s$ heads. 
We concatenate the tensors returned by all heads along the channel axis, obtaining a tensor 
that is further passed through a pointwise convolutional layer reducing the number of activation maps to the number of channels $c$. The tensor $P$ goes through a skip connection, being added to the resulting output, denoted as $Z^*$. 

To significantly improve efficiency, we replace the multi-layer perceptron that typically follows in a conventional transformer block with two pointwise convolutional layers. The first pointwise convolutional layer comprises $4\cdot c$ filters, and the second one only $c$ filters. 

The output of the convolutional transformer is fed into three downscaling heads returning anomaly maps at various resolutions. Each anomaly detection head is formed of two convolutional layers, each with kernels of $3\times 3$ and ReLU activations. The first conv layer is formed of $64$ filters. The second conv layer comprises only one filter, and is followed by a pooling layer which returns the output anomaly map at the desired resolution. At inference time, we aggregate the anomaly maps returned by the three heads into a single frame-level anomaly score by computing the maximum value for each map, and averaging the maximum values.

\noindent
\textbf{Student pre-training.}
Our main objective is to transfer the anomaly detection capabilities from teacher networks to our student model. However, if we start the knowledge distillation with random weights for the student, the network must concurrently learn a high-level representation of the training scenes, and replicate the anomaly detection abilities of the teachers. In this scenario, the objective becomes significantly more difficult to complete. Therefore, we propose to independently address the scene understanding task by training the student as an auto-encoder to reconstruct input video frames, before introducing knowledge distillation. 
More precisely, we train the network to reconstruct the middle frame of a temporal sequence of frames. We fix the length of the temporal sequence to three frames, taken at a stride of $t$. Given an input sequence $\left[x^{i-t}, x^{i},  x^{i+t}\right]$ and the corresponding output of the auto-encoder $\hat{x}^i = D\left(E\left(\left[x^{i-t}, x^{i},  x^{i+t}\right]\right)\right)$, the optimization objective is defined as follows:
\begin{equation}
 \label{eq_ae}
    \begin{split}
        \mathcal{L}_{\mbox{\scriptsize{AE}}} = \frac{1}{h \cdot w\cdot c}\sum_{j=1}^{h}\sum_{k=1}^w\sum_{l=1}^{c}{\left(x^i_{jkl} - \hat{x}^i_{jkl}\right)^2},
    \end{split}
\end{equation}
where $E$ and $D$ represent the encoder and decoder networks. Following \citet{Ionescu-CVPR-2019}, we set $t=3$ as the temporal step for the input sequence. The encoder $E$ is formed of the initial downsampling block and the convolutional transformer (the multi-scale downsampling blocks are added later, during distillation). The decoder $D$ is composed of transposed convolutions, being symmetric with the downsampling block of the encoder.

\noindent
\textbf{Standard knowledge distillation.}
As the first teacher, denoted as $T_1$, we choose the framework based on object-level motion and appearance auto-encoders of \citet{Georgescu-TPAMI-2021}. We select the 3D object-centric network based on self-supervised multi-task learning \citep{Georgescu-CVPR-2021} as the second teacher, denoted as $T_2$. The chosen teacher networks \citep{Georgescu-CVPR-2021,Georgescu-TPAMI-2021} predict the anomaly score of each object detected by a pre-trained detector. Hence, an anomaly map containing the anomaly score associated with each bounding box can be obtained for each video frame. Consequently, our first distillation strategy is to train the student network to regress to the anomaly maps of the teachers, thus learning to imitate the teachers. To make the imitation task more accessible for the student, we lower the resolution of the target anomaly maps by passing them through a max pooling layer. Additionally, we employ multiple resolutions for the anomaly maps, allowing our student model to deal with objects seen at different scales. Let $x_T$ and $x_S$ be the input sequences of frames (centered on some frame $x$) for the teacher and the student. Let $r$ be the number of output resolutions, and $y=T(x_T)$ and $\hat{y}=S(x_S)$ the anomaly maps of some teacher $T$ and the student $S$, respectively. For some teacher $T$, we optimize the following loss function:
\begin{equation}
 \label{pixel_distillation}
    \begin{split}
        \mathcal{L}_{\mbox{\scriptsize{KD}}}(T, S) = \sum_{k=1}^{r}\frac{1}{h_k \cdot w_k}\sum_{i=1}^{h_k}\sum_{j=1}^{w_k}{\left(y^{k}_{ij}-\hat{y}^{k}_{ij}\right)^{2}},
    \end{split}
\end{equation}
where $h_k$ and $w_k$ are the height and the width of the anomaly maps $y^k$ and $\hat{y}^k$ at the $k$-th resolution. In practice, we set the number of output resolutions to $r=3$.

Eq.~\eqref{pixel_distillation} expresses the loss with respect to only one teacher network. However, to increase the generalization power of the student, we distill knowledge from two different anomaly detection models, denoted as $T_1$ and $T_2$. To this end, the objective becomes:
\begin{equation}
 \label{final_pixel}
    \begin{split}
        \mathcal{L}_{\mbox{\scriptsize{KD}}} = \mathcal{L}_{\mbox{\scriptsize{KD}}}(T_1, S) + 
        \mathcal{L}_{\mbox{\scriptsize{KD}}}(T_2, S).
    \end{split}
\end{equation}

\noindent
\textbf{Adversarial distillation.}
Our student is applied directly at the frame level, which means that it needs to learn to jointly detect objects and predict anomalies in a single pass, having a significantly more difficult task than its teachers. Thus, to better mimic the behavior of the teachers, we leverage the use of adversarial training, a procedure typically employed to train generative adversarial networks (GANs) \citep{Goodfellow-NIPS-2014}. For each teacher involved in the distillation, we introduce a discriminator trained to label the anomaly maps produced by the student as fake examples (class $0$) and the ones produced by the teacher as genuine examples (class $1$). As mentioned previously, the student outputs multi-resolution anomaly maps. Each discriminator receives all anomaly map variants and concatenates them (at the appropriate levels) into a single neural network. Therefore, each discriminator makes a single prediction for all anomaly map resolutions.

Let $D_T$ be the discriminator corresponding to a teacher $T$. We train $D_T$ and $S$ through a zero-sum game, alternating between minimizing (for $S$) and maximizing (for $D_T$) the following loss:
\begin{equation}
 \label{adv_distillation}
       \begin{split}
        \mathcal{L}_{\mbox{\scriptsize{AKD}}}(D_T, S) &= \mathbb{E}_{x_T \sim p_T}[\log D_T(T(x_T))] \\ 
        &+ \mathbb{E}_{x_S \sim p_S}[\log 1 - D_T(S(x_S))],
    \end{split}
\end{equation}
where $x_T$ and $x_S$ are input sequences of frames for the teacher and the student models, and $p_T$ and $p_S$ are data densities producing samples for the teacher and the student, respectively. We hereby underline that the teacher model $T$ is frozen. In practice, the data densities $p_T$ and $p_S$ are identical and $x_T=x_S$.

To encapsulate both teachers in the objective formulation, we need to apply Eq.~\eqref{adv_distillation} for each discriminator and sum the corresponding losses, as follows:
\begin{equation}
 \label{final_adv}
    \begin{split}
        \mathcal{L}_{\mbox{\scriptsize{AKD}}} &= \mathcal{L}_{\mbox{\scriptsize{AKD}}}(D_{T_1}, S) + \mathcal{L}_{\mbox{\scriptsize{AKD}}}(D_{T_2}, S),
    \end{split}
\end{equation}
where $D_{T_1}$ and $D_{T_2}$ are the discriminators of teachers $T_1$ and $T_2$, respectively.

\noindent
\textbf{Discriminator architecture.} We employ a convolutional neural network (CNN) as adversarial discriminator, structured into three distinct blocks, as depicted in Figure~\ref{fig_arch}. The initial block processes the anomaly map at the highest resolution, and its output is merged with the anomaly map at the middle resolution to form the input for the subsequent block. Similarly, the output from the second block is combined with the anomaly map at the lowest resolution and fed into the final block. The convolutional blocks share a uniform architecture, namely a convolutional layer succeeded by a max-pooling layer. The first two blocks are based on convolutional layers with $3\times 3$ filters applied at a stride of $1$, using a zero-padding of $1$ pixel in all directions. All units are activated by ReLU. The first conv layer contains $16$ filters and the second is equipped with $32$ filters. The third block is based on a dense layer of $100$ neurons activated by ReLU. A final Softmax layer determines whether the detected anomalies originate from a teacher or a student model.

\noindent
\textbf{Joint distillation.}
The final objective of our framework combines the losses defined in Eq.~\eqref{final_pixel} and Eq.~\eqref{final_adv} into a single loss function as follows:
\begin{equation}
 \label{eq_total}
    \begin{split}
        \mathcal{L}_{\mbox{\scriptsize{total}}} = \mathcal{L}_{\mbox{\scriptsize{KD}}} + \alpha \cdot \mathcal{L}_{\mbox{\scriptsize{AKD}}},
    \end{split}
\end{equation}
where the adversarial loss receives a fixed weight $\alpha=0.1$.

\section{Experiments}
\label{sec_experiments}
\subsection{Data sets}

We conduct experiments on three of the most popular data sets for video anomaly detection: Avenue \citep{Lu-ICCV-2013}, ShanghaiTech \citep{Luo-ICCV-2017} and UCSD Ped2 \citep{Mahadevan-CVPR-2010}.

\noindent{\bf Avenue.}
The Avenue data set \citep{Lu-ICCV-2013} includes 16 training videos (15,328 frames) of normal events, and 21 test videos (15,324 frames) with both normal and abnormal situations. The total number of video frames is 30,652.

\noindent{\bf ShanghaiTech.} The ShanghaiTech data set \citep{Luo-ICCV-2017} is a much larger data set, containing 330 training videos of normal events, and 107 test videos with normal and abnormal scenes. The total number of frames is about 320K, with over 270K being used for training.

\noindent{\bf UCSD Ped2.} UCSD Ped2 \citep{Mahadevan-CVPR-2010} contains 16 training videos and 12 test videos. As the other two data sets, the training videos depict only normal scenarios. 

\subsection{Experimental setup}

\noindent{\bf Teachers.} 
Our frame-level network is trained to replicate the output of two highly effective object-level teachers \citep{Georgescu-TPAMI-2021, Georgescu-CVPR-2021} designed for video anomaly detection. Each teacher detects objects with a pre-trained YOLOv3 \citep{Redmon-arXiv-2018}. The objects are further processed by an object-centric anomaly model. We use the official code to train the teachers\footnote{\url{https://github.com/lilygeorgescu/AED}}$^,$\footnote{\url{https://github.com/lilygeorgescu/AED-SSMTL}}.

\citet{Georgescu-TPAMI-2021} presents a comprehensive framework comprising an object detector, an optical flow estimator, and three auto-encoders, each dedicated to a distinct reconstruction task. Two of these auto-encoders are tasked with reconstructing forward and backward motion, while the third focuses on reconstructing the appearance. The architecture of each encoder is based on three convolutional layers. The structure of each decoder is organized into three blocks, with each block consisting of upsampling and convolution operations, mirroring the encoding architecture. It is important to note that the overall time efficiency of this method, designated as our first teacher, is significantly influenced by the speeds of the object detector and the optical flow estimator, respectively.

\citet{Georgescu-CVPR-2021} proposes an object-centric method, which we choose to serve as the second teacher for our student net. The detected objects are fed into a 3D CNN that serves as a feature extractor, followed by four prediction modules. Three of these modules utilize a simple architecture, comprising one convolutional layer followed by a fully-connected layer. The fourth module, designed to estimate the masked central bounding box in an object-centric sequence, mirrors the structure of the 3D CNN, but employs 2D convolutional layers to accommodate its specific predictive task. The authors explore several configurations of the 3D CNN, with the most comprehensive and effective version featuring nine convolutional layers. The width of these layers varies, with the number of filters ranging from 32 to 64.

We underline that our frame-level student uses a transformer-based architecture, while the teachers are based on complex frameworks comprising multiple networks. For instance, the teachers employ a YOLO-based model to detect objects, and convolutional architectures to label objects with anomaly scores. In contrast, our student uses a single end-to-end network, thus having a simplified and efficient structure.

\begin{table*}[th!]
\centering 
\setlength\tabcolsep{3.2pt}
\small{
\begin{tabular}{| c | l | c | c | c | c | c | c | c |} 
\hline
 \multirow{2}{*}{\rotatebox[origin=c]{90}{Type}} &
 \multirow{2}{*}{Method} & \multicolumn{2}{c|}{CHUK Avenue} & \multicolumn{2}{c|}{ShanghaiTech} & \multicolumn{2}{c|}{UCSD Ped2} & \multirow{2}{*}{FPS}\\
 \cline{3-8}
 & & \rotatebox[origin=c]{0}{Micro AUC} & \rotatebox[origin=c]{0}{Macro AUC} & \rotatebox[origin=c]{0}{Micro AUC} & \rotatebox[origin=c]{0}{Macro AUC} & \rotatebox[origin=c]{0}{Micro AUC} & \rotatebox[origin=c]{0}{Macro AUC} & \\
 \hline 
 \hline
 {\multirow{13}{*}{\rotatebox[origin=c]{90}{Object-centric}}} 
&
\cite{Barbalau-CVIU-2023} & 91.6 & 92.5 & {\color{ForestGreen}83.8} & {\color{ForestGreen}90.5} & - & - & 20\\
 & \cite{Doshi-CVPRW-2020a}  & 86.4 & - & 71.6 & - & {97.8} & - & -\\
& 
\cite{Georgescu-CVPR-2021} &  91.5 & {\color{RoyalBlue}92.8} & 82.4 & {\color{RoyalBlue}90.2} & 97.5 & {\color{red}99.8} & 51 \\
& 
\cite{Georgescu-TPAMI-2021} &  {\color{RoyalBlue}92.3} & 90.4 &  82.7 & 89.3 & {\color{RoyalBlue}98.7} & {\color{ForestGreen}99.7} & 24 \\
 &
 \cite{Ionescu-CVPR-2019} & 87.4 & 90.4 & 78.7 & 84.9 & 94.3 & {\color{RoyalBlue}97.8}  & -\\
&
\cite{Liu-ICCV-2021} & 89.9 & {\color{ForestGreen}93.5} & 74.2 & 83.2 & {\color{red}99.3} & -& 12 \\ 
&
\cite{Madan-ARXIV-2022} + \cite{Barbalau-CVIU-2023} & 91.6 & {92.4} &  {\color{RoyalBlue}83.6} & {\color{red}90.6} & - & - & 20\\
&
\cite{Madan-ARXIV-2022} + \cite{Georgescu-TPAMI-2021} & {\color{red}93.2} & 91.8 & 83.3 & 89.3 & - & - & 31 \\
&
\cite{Madan-ARXIV-2022} + \cite{Liu-ICCV-2021} &  89.5 & {\color{red}93.6} &  75.2 & 83.8 & - & - & 10\\ 
&
\cite{Ristea-CVPR-2022} + \cite{Georgescu-TPAMI-2021} & {\color{ForestGreen}92.9} & 91.9 & {\color{RoyalBlue}83.6} & 89.5 & - &- & 31 \\
&
\cite{Ristea-CVPR-2022} + \cite{Liu-ICCV-2021} &  {90.9} & {92.2} & 75.5 & {83.7} & - & - & 10\\ 
 &
 \cite{Wang-ECCV-2022} & 92.2 & - & {\color{red}84.3} & - & {\color{ForestGreen}99.0} & - & 35\\
 &
 \cite{Yu-ACMMM-2020} & 89.6 & - & 74.8 & - & 97.3 & - & -\\

\hline
{\multirow{28}{*}{\rotatebox[origin=c]{90}{Frame or cube level}}}

&
\cite{Astrid-BMVC-2021} & 87.1 & - & 75.9 & - & 96.5 & - &-\\
&
\cite{Astrid-ICCVW-2021} & 84.7 & - & 73.7 & - & {\color{red}98.4}&- &-\\
&
\cite{Gong-ICCV-2019} & 83.3 & -  & 71.2 & - & 94.1 & - & 35\\
&
\cite{Ionescu-WACV-2019} & 88.9 & - & - & - & - & - &- \\
&
\cite{Lee-TIP-2019} & {\color{ForestGreen}90.0} & - &- & - &96.6 & - &-\\
&
\cite{Liu-CVPR-2018} & 85.1 & 81.7 & 72.8 & 80.6 & 95.4& - & 28\\
&
\cite{Liu-BMVC-2018} & 84.4 & - & - & - & 87.5 & - &-\\
&
\cite{Madan-ARXIV-2022} + \cite{Liu-CVPR-2018} & 89.1 & 84.8 & 74.6 & {\color{ForestGreen}83.3} & - & - & 26\\
&
\cite{Madan-ARXIV-2022} + \cite{Park-CVPR-2020} & 86.4 & 86.3 & 70.6 & 80.3 & - & - & 94\\ 
&
\cite{Nguyen-ICCV-2019} & 86.9 & - &  - & - & 96.2 & - &-\\
&
\cite{Park-WACV-2022} & 85.3 & - & 72.2 & - & 96.3 & - & 195\\
&
\cite{Park-CVPR-2020} & 82.8 & {\color{RoyalBlue}86.8} & 68.3 & 79.7& {\color{RoyalBlue}97.0} & - & 101\\
&
\cite{Ramachandra-WACV-2020a} & 72.0 & - &  - & - & 88.3 & - &-\\
&
\cite{Ramachandra-WACV-2020b} & 87.2 & - & - & - & 93.0 & - &-\\
&
\cite{Ravanbakhsh-ICIP-2017} & - & - & - & - & 93.5 & - &-\\
&
\cite{Ravanbakhsh-WACV-2018} & - & - & - & - & 88.4 & - &-\\
&
\cite{Ristea-CVPR-2022} + \cite{Liu-CVPR-2018} & 87.3 & 84.5 & 74.5 & {\color{RoyalBlue}82.9} & - & - & 26\\
&
\cite{Ristea-CVPR-2022} + \cite{Park-CVPR-2020} & 84.8 & {\color{red}88.6} & 69.8 & 80.2 & - & - & 95 \\
&
\cite{Smeureanu-ICIAP-2017} & 84.6 & - & - & - & - & - & -\\
&
\cite{Sultani-CVPR-2018} & - & - & - & 76.5 & - & - & 56 \\
&
\cite{Sun-ACMMM-2020}  & {\color{RoyalBlue}89.6} & - & 74.7 & - & - & - & -\\
&
\cite{Tang-PRL-2020} & 85.1 & - & 73.0 & - & 96.3 & - & -\\
&
\cite{Wang-ACMMM-2020} & 87.0 & - & {\color{ForestGreen}79.3} & - & - & - & -\\
&
\cite{Wu-ECCV-2022} &  - & - & {\color{red}80.4} & - & - & - & -\\
&
\cite{Wu-TNNLS-2019} & 86.6 & - & - & - & 96.9 & -  & -\\
&
\cite{Yu-TNNLS-2021} & {\color{red}90.2} & - & - & - & {\color{ForestGreen}97.3} & - & -\\
&
\cite{Zhang-PR-2016} & - & - & - & - & 91.0 & - & - \\
\cline{2-9}
&Ours & 88.3 &  {\color{ForestGreen}87.9} & {\color{RoyalBlue}78.4} & {\color{red}86.0} & 95.9 & 97.1 & 1480\\
\hline
\end{tabular}
}
\caption{Micro and macro AUC scores (in \%) of our method versus several state-of-the-art frame-level and object-level methods on Avenue, ShanghaiTech and UCSD Ped2. For each family of methods, the best score is shown in {\color{red}red}, the second-best in {\color{ForestGreen}green}, and third-best in {\color{RoyalBlue}blue}. The reported FPS values (including those of the baselines) are measured on a machine with an Nvidia GeForce GTX 3090 GPU with 24 GB of VRAM.}
\label{tab_results} 
\end{table*}



\noindent{\bf Baselines.} 
We compare our method with state-of-the-art frame-level \citep{Astrid-BMVC-2021, Astrid-ICCVW-2021, Gong-ICCV-2019, Ionescu-WACV-2019, Liu-BMVC-2018, Liu-CVPR-2018, Liu-ICCV-2021, Madan-ARXIV-2022, Nguyen-ICCV-2019, Park-CVPR-2020, Ramachandra-WACV-2020a, Ramachandra-WACV-2020b, Ravanbakhsh-WACV-2018, Ravanbakhsh-ICIP-2017, Ristea-CVPR-2022, Smeureanu-ICIAP-2017, Sultani-CVPR-2018, Sun-ACMMM-2020, Tang-PRL-2020, Wang-ACMMM-2020, Wu-TNNLS-2019, Wu-ECCV-2022, Yu-TNNLS-2021, Zhang-PR-2016} and object-level \citep{Barbalau-CVIU-2023, Doshi-CVPRW-2020a,Georgescu-CVPR-2021,Georgescu-TPAMI-2021,Ionescu-CVPR-2019,Madan-ARXIV-2022,Ristea-CVPR-2022, Wang-ECCV-2022,Yu-ACMMM-2020} frameworks.

\noindent{\bf Hyperparameters.} We opt for a lightweight model having under 5M parameters (as reference, ResNet-18 has 11M parameters), setting the number of transformer blocks to $m\!=\!5$, the number of self-attention heads to $s\!=\!5$, 
and the number of output heads to $3$. For Avenue and ShanghaiTech, the output resolutions are $1\times1$, $4\times 4$ and $16\times 16$. Since the objects in UCSD Ped2 have a much smaller scale, the output resolutions are set to $1\times1$, $30\times 30$ and $60\times 60$. Regardless of the data set, the student network is trained for $35$ epochs on mini-batches of $64$ video sequences of $3$ frames each. We optimize the model with Adam \citep{Kingma-ICLR-2014}, using a learning rate of $10^{-4}$ and a weight decay set to $10^{-5}$ (all other hyperparameters are set to default values).

\noindent{\bf Evaluation.} Following recent works \citep{Acsintoae-CVPR-2022,Georgescu-TPAMI-2021,Ristea-CVPR-2022}, we evaluate detection performance in terms of the micro-averaged and macro-averaged frame-level AUC, and localization performance in terms of the Region-Based Detection Criterion (RBDC) and Track-Based Detection Criterion (TBDC). For both AUC measures, we compute the area under the ROC curve (AUC) with respect to the ground-truth frame-level annotations. A frame is labeled as abnormal if the predicted anomaly score is larger than the current threshold. For the micro AUC, all the test frames are first concatenated into a single video before computing the AUC. For the macro AUC, we compute the AUC of each test video and report the mean of the resulting AUC scores. As recommended by \citet{Ramachandra-WACV-2020a}, we set the region overlap threshold to $0.1$ for RBDC, and the track overlap threshold to $0.1$ for TBDC.

\begin{table*}[t]
\centering 
\small{
\begin{tabular}{| c | l | c | c | c | c | c |} 
\hline
 \multirow{2}{*}{{Type}} &
 \multirow{2}{*}{Method} & \multicolumn{2}{c|}{Avenue} & \multicolumn{2}{c|}{Shanghai} &
 \multirow{2}{*}{FPS} \\
 \cline{3-6}
 & & {RBDC} & {TBDC} & {RBDC} & {TBDC} & \\
 \hline 
 \hline
 
 {\multirow{10}{*}{{Object-centric}}} 
&
\cite{Barbalau-CVIU-2023} & 47.83 & 85.26 & {\color{ForestGreen}47.14} & {\color{ForestGreen}85.61} & 20\\
& 
\cite{Georgescu-CVPR-2021} &  57.00 & 58.30 & 42.80 & 83.90 & 51 \\
& 
\cite{Georgescu-TPAMI-2021} & {\color{RoyalBlue}65.05} & 66.85 & 41.34 & 78.79 & 24 \\
&
 \cite{Ionescu-CVPR-2019} & 15.77 & 27.01 & 20.65 & 44.54 & -\\
 &
\cite{Liu-ICCV-2021} & 41.05 & {\color{RoyalBlue}86.18} & 44.41 & 83.86 & 12 \\ 
&
\cite{Madan-ARXIV-2022} + \cite{Barbalau-CVIU-2023} & 49.01 & 85.94 &  {\color{red}47.73} & {\color{red}85.68} & 20\\
&
\cite{Madan-ARXIV-2022} + \cite{Georgescu-TPAMI-2021} & {\color{red}66.04} & 65.12 & 40.52 & 81.93 & 31\\
&
\cite{Madan-ARXIV-2022} + \cite{Liu-ICCV-2021} & 46.49 & {\color{ForestGreen}86.43} & 45.86 &  {\color{RoyalBlue}84.69} & 10 \\ 
&
\cite{Ristea-CVPR-2022} + \cite{Georgescu-TPAMI-2021} & {\color{ForestGreen}65.99} & 64.91 & 40.55 & 83.46 & 31 \\
&
\cite{Ristea-CVPR-2022} + \cite{Liu-ICCV-2021} & 62.27 & {\color{red}89.28} &  {\color{RoyalBlue}45.45} & 84.50 & 10\\ 

\hline
{\multirow{5}{*}{{Frame or}}}
&
\cite{Liu-CVPR-2018} & 19.59 & 56.01 & 17.03 & 54.23 & 28\\
{\multirow{5}{*}{{cube level}}} &
\cite{Madan-ARXIV-2022} + \cite{Liu-CVPR-2018} & 23.79 & {\color{RoyalBlue}66.03} & {\color{ForestGreen}19.13} & {\color{ForestGreen}61.65} & 26\\
&
\cite{Ramachandra-WACV-2020a} & {\color{RoyalBlue}35.80} & {\color{red}80.90} & - & - & -\\
&
\cite{Ramachandra-WACV-2020b} & {\color{ForestGreen}41.20} & {\color{ForestGreen}78.60} & - & - & -\\
&
\cite{Ristea-CVPR-2022} + \cite{Liu-CVPR-2018} & 20.13 & 62.30 & {\color{RoyalBlue}18.51} & {\color{RoyalBlue}60.22} & 26\\
\cline{2-7}
& Ours & {\color{red}41.92} & 44.49  & {\color{red}19.34} & {\color{red}63.64} & 1480\\
\hline
\end{tabular}
}
\caption{RBDC and TBDC scores (in \%) of our method versus several state-of-the-art frame-level and object-level methods on Avenue and ShanghaiTech. For each family of methods, the best score is shown in {\color{red}red}, the second-best in {\color{ForestGreen}green}, and third-best in {\color{RoyalBlue}blue}. The reported FPS values (including those of the baselines) are measured on a machine with an Nvidia GeForce GTX 3090 GPU with 24 GB of VRAM.}
\label{tab_results_rbdc_tbdc} 
\end{table*}

\subsection{Results}

\subsubsection{Quantitative analysis}
In Table~\ref{tab_results}, we present the micro and macro AUC scores of our method in comparison with the state-of-the-art frame-level and object-level methods on the Avenue, ShanghaiTech and UCSD Ped2 benchmarks. Even if our method is below the top scoring object-centric methods, it yields comparable performance with some methods from this category, \eg~\citep{Ionescu-CVPR-2019,Yu-ACMMM-2020}. Moreover, with a micro AUC of $88.3\%$, our student outperforms most of the frame-based models. Among all frame-level frameworks reporting results on ShanghaiTech, we observe that our model ranks first in terms of the macro AUC, and third in terms of the micro AUC. With respect to the teachers \citep{Georgescu-CVPR-2021,Georgescu-TPAMI-2021}, our micro and macro AUC drops are lower than $5\%$. In terms of the macro AUC on UCSD Ped2, our model is just below the state-of-the-art object-centric approach \citep{Georgescu-TPAMI-2021}, at a difference of only $2.7\%$. In terms of the micro AUC, the student registers a performance drop of at most $2.8\%$ with respect to both teachers \citep{Georgescu-CVPR-2021,Georgescu-TPAMI-2021}. 

In Table \ref{tab_results_rbdc_tbdc}, we compare the RBDC and TBDC scores of our method with the RBDC and TBDC scores of other studies which reported these localization metrics. In general, frame-level methods are less able to localize anomalies than object-centric methods. This explains why the number of frame-level methods reporting the RBDC and TBDC scores is much lower than the number of object-centric methods. Notably, our method reaches the highest RBDC scores among all frame-level methods, on both Avenue and ShanghaiTech. While our TBDC score is not competitive on the Avenue data set, we attain the highest TBDC score on ShanghaiTech, when our method is compared with methods from the same category. Furthermore, we observe that our method also surpasses the object-centric method of \citet{Ionescu-CVPR-2019}. We conclude that the localization performance of our approach is competitive. 

\subsubsection{Qualitative analysis}
\begin{figure}[!t]
\begin{center}
\centerline{\includegraphics[width=1.0\linewidth]{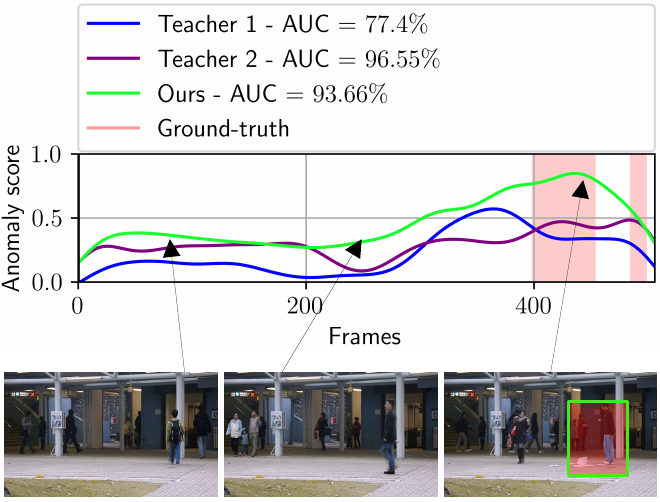}}
\vspace{-0.25cm}
\caption{Comparing the frame-level anomaly scores of teachers $T_1$ \citep{Georgescu-TPAMI-2021} and $T_2$ \citep{Georgescu-CVPR-2021} with the scores of our student on test video {14} from Avenue. The anomaly localization examples are provided by the head with the highest resolution of our student. Best viewed in color.}
\label{fig_avenue}
\end{center}
\end{figure}

\begin{figure}[!t]
\begin{center}
\centerline{\includegraphics[width=1.0\linewidth]{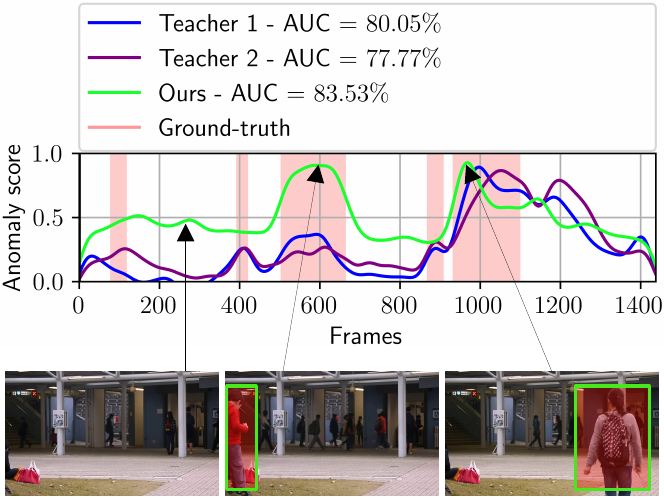}}
\vspace{-0.25cm}
\caption{Comparing the frame-level anomaly scores of teachers $T_1$ \citep{Georgescu-TPAMI-2021} and $T_2$ \citep{Georgescu-CVPR-2021} with the scores of our student on test video {1} from Avenue. The anomaly localization examples are provided by the head with the highest resolution of our student. Best viewed in color.}
\label{fig_avenue_supp}
\end{center}
\end{figure}
\begin{figure}[!t]
\begin{center}
\centerline{\includegraphics[width=1.0\linewidth]{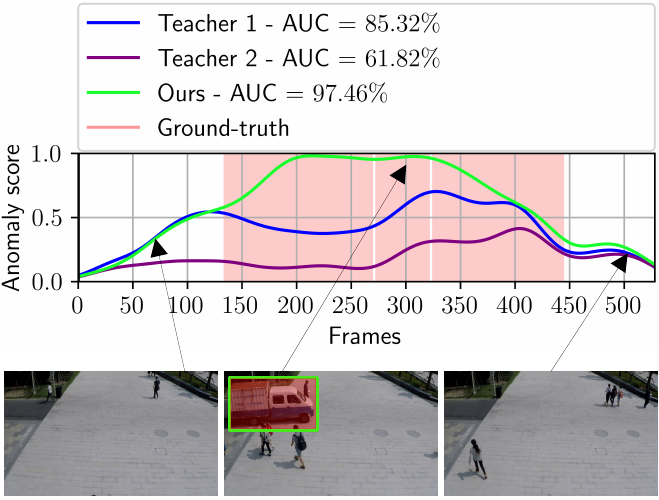}}
\vspace{-0.25cm}
\caption{Comparing the frame-level anomaly scores of teachers $T_1$ \citep{Georgescu-TPAMI-2021} and $T_2$ \citep{Georgescu-CVPR-2021} with the scores of our student on test video {01\_0136} from ShanghaiTech. The anomaly localization examples are provided by the head with the highest resolution of our student. Best viewed in color.}
\label{fig_shanghai_supp}
\end{center}
\end{figure}

\begin{figure}[!t]
\begin{center}
\centerline{\includegraphics[width=1.0\linewidth]{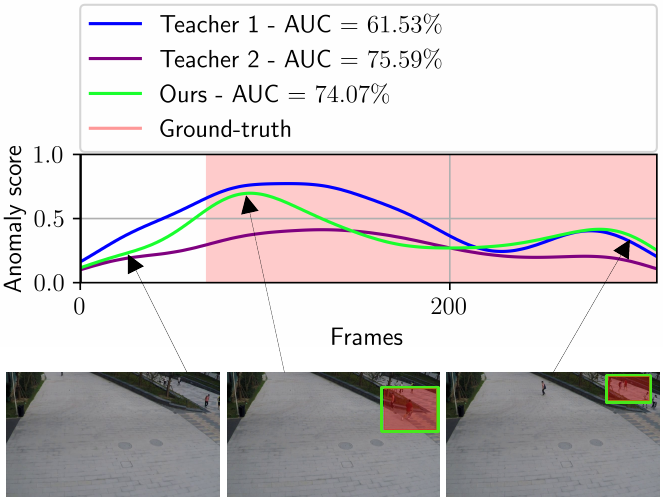}}
\vspace{-0.25cm}
\caption{Comparing the frame-level anomaly scores of teachers $T_1$ \citep{Georgescu-TPAMI-2021} and $T_2$ \citep{Georgescu-CVPR-2021} with the scores of our student on test video 01\_0055 from ShanghaiTech. The anomaly localization examples are provided by the head with the highest resolution of our student. Best viewed in color.}
\label{fig_shanghai}
\end{center}
\end{figure}

\begin{figure}[!t]
\begin{center}
\centerline{\includegraphics[width=1.0\linewidth]{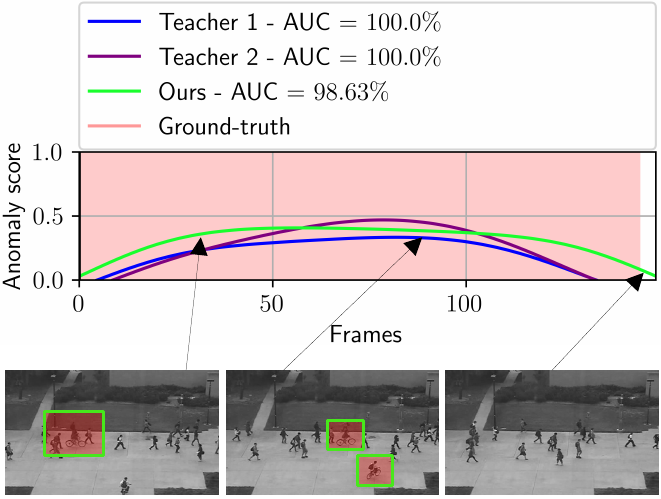}}
\vspace{-0.25cm}
\caption{Comparing the frame-level anomaly scores of teachers $T_1$ \citep{Georgescu-TPAMI-2021} and $T_2$ \citep{Georgescu-CVPR-2021} with the scores of our student on test video {3} from UCSD Ped 2. The anomaly localization examples are provided by the head with the highest resolution of our student. Best viewed in color.}
\label{fig_ped2_supp}
\end{center}
\end{figure}

\begin{figure}[!t]
\begin{center}
\centerline{\includegraphics[width=1.0\linewidth]{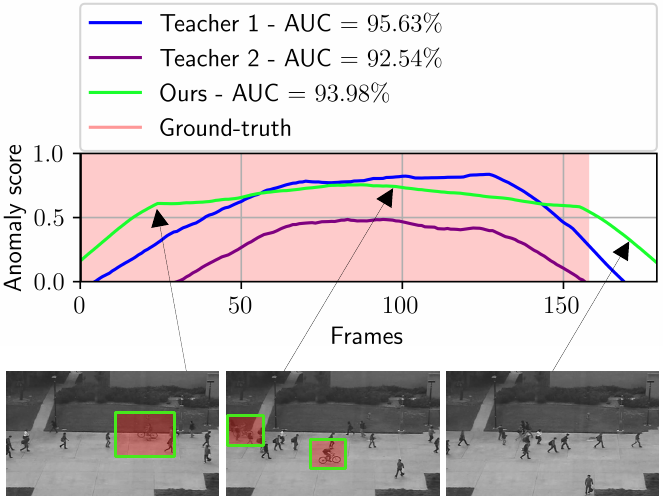}}
\vspace{-0.25cm}
\caption{Comparing the frame-level anomaly scores of teachers $T_1$ \citep{Georgescu-TPAMI-2021} and $T_2$ \citep{Georgescu-CVPR-2021} with the scores of our student on test video {6} from UCSD Ped2. The anomaly localization examples are produced by the head with the highest resolution of our student. Best viewed in color.}
\label{fig_ped2}
\end{center}
\end{figure}

\noindent{\bf Frame-level scores.}
We provide 
frame-level output visualizations on multiple test videos. In Figure~\ref{fig_avenue}, we illustrate the anomaly detection performance on test video 14 from Avenue. Our student is able to surpass the lower results of teacher $T_1$, being very close to teacher $T_2$. This shows the importance of learning from more than one teacher. In Figure~\ref{fig_avenue_supp}, we present the anomaly detection performance on test video 1 from Avenue. On this video, our student is remarkably able to surpass both teachers by more than $3\%$ in terms of the AUC. This highlights that, even if the complexity of the student network is significantly lower compared with the complexity of the teachers, the student is able to learn aggregated information, which can eventually lead to surpassing the teachers in certain cases. An even higher gap in favor of our student is illustrated in Figure~\ref{fig_shanghai_supp}. Indeed, the frame-level anomaly predictions on test video {01\_0136} from ShanghaiTech indicate that our student surpasses the best teacher, $T_1$ \citep{Georgescu-TPAMI-2021}, by more than $12\%$, attaining an overall AUC of $97.46\%$. In Figure~\ref{fig_shanghai}, we showcase the anomaly detection performance on test video 01\_0055 from ShanghaiTech. Our student, being trained to mimic both teachers, is able to surpass the lower results of teacher $T_1$, getting close to the performance of teacher $T_2$. For the UCSD Ped2 data set, we showcase the frame-level scores on test video 3 in Figure~\ref{fig_ped2_supp}. In this example, the student model attains an AUC score of $98.63\%$, being very close to the teachers, which have perfect scores. In Figure~\ref{fig_ped2}, we present the anomaly detection performance on test video 6 from UCSD Ped2. Here, the performance of our student is between that of the teachers $T_1$ and $T_2$. 

\begin{figure*}[!th]
\begin{center}
\centerline{\includegraphics[width=1.0\linewidth]{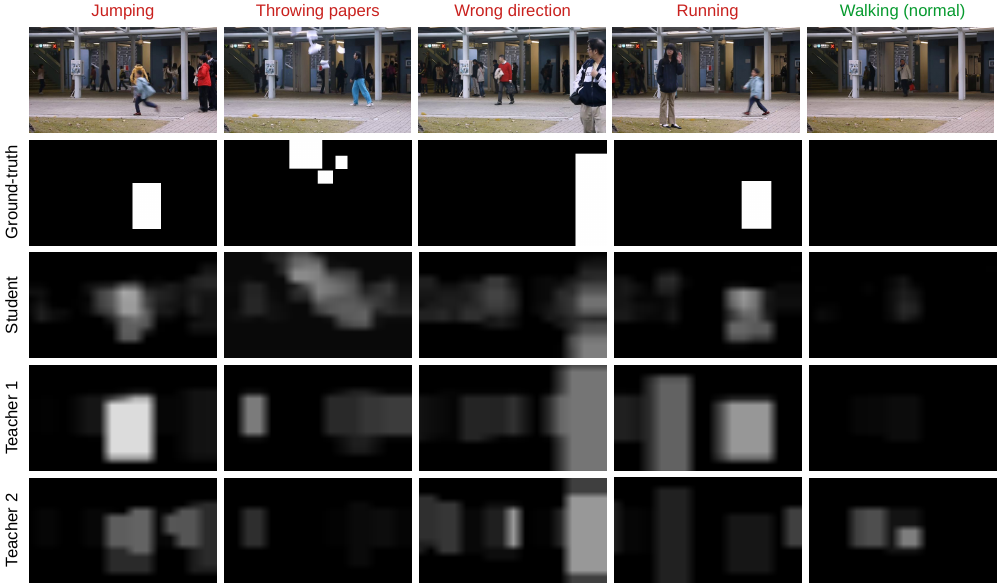}}
\vspace{-0.25cm}
\caption{Examples of anomaly maps of $16\times 16$ pixels generated by our student and the two teachers, for a set of five randomly chosen test frames from Avenue. For reference, the ground-truth anomaly maps from Avenue are also included. Best viewed in color.}
\label{fig_anomaly_maps}
\end{center}
\end{figure*}

\begin{figure}[!th]
\begin{center}
\centerline{\includegraphics[width=1.0\linewidth]{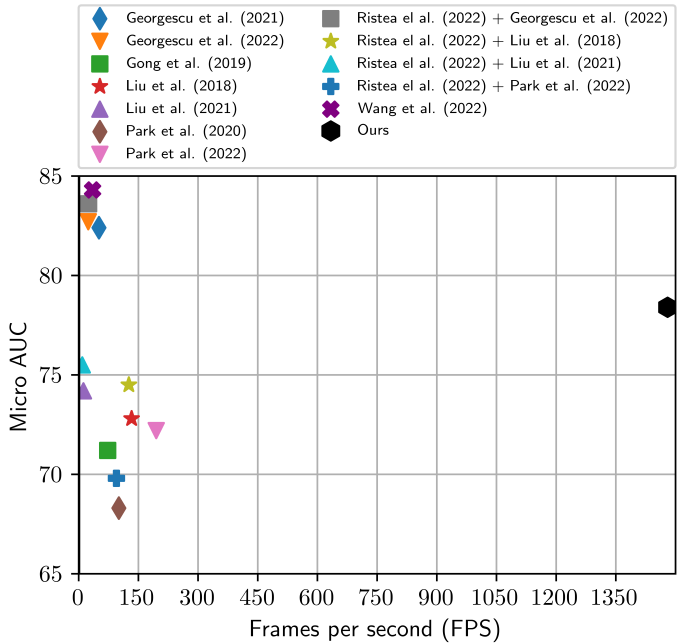}}
\vspace{-0.25cm}
\caption{The trade-off between performance (micro AUC) and speed (FPS) for our student versus multiple state-of-the-art methods \citep{Georgescu-CVPR-2021,Georgescu-TPAMI-2021,Gong-ICCV-2019,Liu-CVPR-2018,Liu-ICCV-2021,Park-CVPR-2020,Park-WACV-2022,Ristea-CVPR-2022,Wang-ECCV-2022} (with open-sourced code), on the ShanghaiTech data set. The running times of all methods are measured on a machine with an Nvidia GeForce GTX 3090 GPU with 24 GB of VRAM. Best viewed in color.}
\label{fig_tradeoff_sh}
\end{center}
\end{figure}

\noindent{\bf Anomaly maps.}
In Figure \ref{fig_anomaly_maps}, we illustrate a set of predicted versus ground-truth anomaly maps for five test frames from Avenue. The anomaly maps on the third row are predicted by our student model, while those on the fourth and fifth rows are predicted by the teachers $T_1$ and $T_2$, respectively. We underline that the anomaly maps generated by the student are smoother, because it does not rely on a pre-trained detector to obtain object bounding boxes. This feature seems to be extremely helpful for the example depicted on the second column, where the teachers fail to detect the papers thrown in the air, since \emph{paper} is not among the classes known by YOLOv3 \citep{Redmon-arXiv-2018}. In contrast, our student assigns high anomaly scores to the region where there are papers in the air. Another interesting behavior for our student is its ability to assign a high score to the child running in the example shown on the fourth column, without triggering high anomaly scores for the person standing on the grass, as opposed to the teachers. The example shown on the last column depicts a normal event, where people are just walking. Here, we observe that teacher $T_2$ outputs high anomaly scores without a good reason. In the end, we believe that one of the most remarkable abilities of our student is to detect anomalies where object-centric approaches based on detectors with a limited number of known classes fail.

\subsection{Speed versus accuracy trade-off}

In Figure \ref{fig_tradeoff_avenue}, we compare our model with several recent frame-level and object-centric methods in terms of the trade-off between accuracy and speed, on the Avenue benchmark. The running times of all methods \citep{Georgescu-CVPR-2021,Georgescu-TPAMI-2021,Gong-ICCV-2019,Ionescu-ICCV-2017,Liu-CVPR-2018,Liu-ICCV-2021,Park-CVPR-2020,Park-WACV-2022,Ristea-CVPR-2022,Wang-ECCV-2022}, including our own, are measured on a machine with an Nvidia GeForce GTX 3090 GPU with 24 GB of VRAM. 
The FPS rates of the object-level methods range between 20 and 52. Some frame-level methods are comparatively faster, the most efficient one, proposed by \citet{Park-WACV-2022}, running at 195 FPS. Our method runs at 1480 FPS, being more than 7 times faster than the most efficient competitor \citep{Park-WACV-2022}. Remarkably, we also outperform the fastest competitor \citep{Park-WACV-2022} in terms of the micro AUC (by $3\%$). 
In Figure~\ref{fig_tradeoff_sh}, we present the same trade-off between accuracy and time on the ShanghaiTech benchmark, and we see that our method outperforms again the fastest competitor, exhibiting a superior micro AUC advantage of $5.8\%$.
 
\noindent{\bf Comparison with FastAno.}
The trade-offs reported on both Avenue and ShanghaiTech clearly indicate FastAno \citep{Park-WACV-2022} as the fastest competing model. On both data sets, our model obtains a better trade-off between speed and accuracy, being more than 7 times faster while reaching scores that are more than $5\%$ better. The state-of-the-art trade-off of our model stems from our option to use a very lightweight architecture (having under 5 million parameters) that is applied on {\bf whole frames}, using a downsampling block to quickly downscale the input resolution. FastAno \citep{Park-WACV-2022} and most frame-level works divide the frames into patches or cuboids, which need to be processed as individual samples in mini-batches. Our results show that processing a single mini-batch of frames (with an aggressive downsampling block) instead of many mini-batches of patches is significantly more efficient and effective. This is made possible by our training procedure based on distilling knowledge from highly accurate object-centric teachers.

\begin{table}[t]
\centering 
\small{
\begin{tabular}{| l | c | c |} 
\hline
   {Usage Parameter} & Teacher $T_2$ & Student  \\
 \hline
  \hline
  Memory usage (GB) & 17 & 6.5 \\
  Trainable parameters (M) & 65 & 5 \\
  GFLOPs & 107.9 & 6.2 \\
 \hline
\end{tabular}
}
\caption{Comparing our student with the more efficient teacher $T_2$ \citep{Georgescu-CVPR-2021} in terms of various usage parameters.}
\label{tab_other_stats} 
\end{table}

\noindent{\bf Comparison with teachers.}
We showed that our student model runs more efficiently on GPU compared with other models. To confirm that the speed gains are not specific to GPU environments, we tested our model and its teachers, $T_1$ and $T_2$, in a different environment (Intel i9 CPU 3.7 GHz, no GPU). On CPU, our model runs at 88 FPS, while the teachers $T_1$ and $T_2$ run at 2 and 4 FPS, respectively. These running times confirm the speed superiority of our model, which is still 22 to 44 times faster than the object-centric teachers. Remarkably, our model can process about 3 video streams at 30 FPS on CPU, while the teachers must rely on the GPU environment to deliver real-time processing for 1 or 2 streams. Thus, our student can process more streams on CPU than its teachers on GPU.

Aside from comparing our model with the teachers in terms of the FPS, we also look at other characteristics, such the memory usage, the number of parameters and the number of floating point operations (GFLOPs). We choose to compare our student with teacher $T_2$ \citep{Georgescu-CVPR-2021}, since this teacher proved to be faster than teacher $T_1$ \citep{Georgescu-TPAMI-2021} in both GPU and CPU environments. The comparison is shown in Table \ref{tab_other_stats}. Considering the additional usage parameters, we find that our student preserves its efficiency, consuming about one third of the RAM and using 17 times less GFLOPs. This is mainly due to our lighter architecture, which has 13 times less trainable weights.



\begin{table}[t]
\centering 
\small{
\begin{tabular}{| c | l | c | c | c |} 
\hline
 \multirow{2}{*}{$m$} &
 \multirow{2}{*}{$s$} & \multicolumn{2}{c|}{AUC} & \multirow{2}{*}{FPS}\\
 \cline{3-4}
 & & Micro & Macro &   \\
 \hline
\hline
 3 & 5 & 67.9 & 72.2 & 2266\\
 4 & 5 & 72.1 & 76.7 & 1806\\
 5 & 5 & 75.4 & 75.5 & 1480\\
 6 & 5 & 68.6 & 77.4 & 1256\\
 7 & 5 & 72.6 & 71.9 & 1100\\
 \hline
 5 & 3 & 70.6 & 73.3 & 1476\\
 5 & 4 & 74.3 & 72.9 & 1487\\
 5 & 5 & 75.4 & 75.5 & 1480\\
 5 & 6 & 68.5 & 70.6 & 1472\\
 5 & 7 & 72.5 & 74.6 & 1460\\
 \hline
\end{tabular}
}
\caption{Micro and macro AUC scores (in \%) and FPS rates on the Avenue data set \citep{Lu-ICCV-2013}, while varying the number of transformer blocks $m$ and the number heads $s$ inside our architecture.}
\label{tab_blocks} 
\end{table}

\subsection{Ablation studies}

We start developing our model from a truncated architecture that takes a single frame as input and produces anomaly maps of $4\times 4$ pixels. The truncated architecture is trained using only standard knowledge distillation from a single teacher, namely $T_1$ \citep{Georgescu-TPAMI-2021}.

We report both micro and macro AUC scores for our ablation studies. When making our design choices, we refer to the micro AUC as the more representative measure, due to its wider popularity (in contrast to the macro AUC) across video anomaly detection papers \citep{Ramachandra-PAMI-2020}.

\noindent{\bf Transformer architecture.} To determine the optimal depth for our convolutional transformer, we first fix the number of heads to $s=5$ and experiment with architectures comprising $m$ blocks, where $m \in \{3,4,5,6,7\}$. The results reported in Table \ref{tab_blocks} indicate that using $m=5$ transformer blocks leads to the highest micro AUC score, while keeping an astonishing FPS rate of 1480. Next, we fix the number of blocks to $m=5$ and variate the number of heads, considering $s \in \{3,4,5,6,7\}$. Unlike the number of blocks, we observe that the number of heads does not have a high impact on the running time. We obtain the highest micro AUC with $s=5$ heads. We underline that all configurations are much faster than competing anomaly detection methods. We hereby focus on choosing the configuration providing the highest micro AUC for the subsequent experiments, namely the one with $m=5$ blocks and $s=5$ heads.

\begin{table}[t]
\centering 
\small{
\begin{tabular}{| c | c | c | c |} 
\hline
 \#Output &
 \multirow{2}{*}{Head Resolution} & \multicolumn{2}{c|}{AUC}\\
 \cline{3-4}
 Heads & & Micro & Macro  \\
 \hline
  \hline
 1 & $1\times1$ & 68.2 & 74.3\\
 1 & $4\times4$ & 75.4 & 75.5\\
 1 & $16\times16$ & 65.7 & 74.0\\
 1 & $64\times64$ & 62.3 & 65.8\\
 1 & Full & 59.2 &	73.9 \\
 \hline
 2 & $1\times1$, $4\times4$ & 71.7 & 74.7 \\
 3 & $1\times1$, $4\times4$, $16\times16$ & \textbf{76.4} & \textbf{76.0}\\
 \hline
\end{tabular}
}
\caption{Micro and macro AUC scores (in \%) on the Avenue data set \citep{Lu-ICCV-2013}, while varying the resolution and the number of output heads. The top scores are highlighted in bold.}
\label{tab_heads} 
\end{table}

\noindent{\bf Output heads.} The task of replicating the full-resolution anomaly maps produced by the state-of-the-art teacher models \citep{Georgescu-CVPR-2021, Georgescu-TPAMI-2021} is quite difficult, and the anomaly detection performance is directly impacted by the ability of the student to replicate the teachers. To simplify the task, we reduce the resolution of the anomaly maps. To determine the optimal resolution, we test different output head configurations, reporting the results in Table \ref{tab_heads}. According to the presented results, our model seems to achieve better micro AUC scores when producing anomaly maps of $1\times 1$ and $4 \times 4$ pixels. The micro AUC tends to drop as we increase the resolution of the anomaly maps. Aside from testing individual output heads, we also consider combining the low-resolution heads into a single architecture with multi-resolution outputs. This approach proves to be useful when combining the $1\times 1$, $4 \times 4$ and $16 \times 16$ output heads. We continue our experiments with the architecture comprising these multi-resolution output heads.

\begin{table}[t]
\centering 
\small{
\begin{tabular}{| c | c | c |} 
\hline
   \multirow{2}{*}{\#Input Frames} &  \multicolumn{2}{c|}{AUC}\\
 \cline{2-3}
  & Micro & Macro  \\
 \hline
  \hline
  1 & 76.4 & 76.0 \\
  3 & \textbf{79.5} & \textbf{80.7} \\
  5 & 78.4 & 80.3 \\
 \hline
\end{tabular}
}
\caption{Micro and macro AUC scores (in \%) on the Avenue data set \citep{Lu-ICCV-2013}, while varying the number of input frames. The top scores are highlighted in bold.}
\label{tab_frames} 
\end{table}

\noindent{\bf Input frames.}
Many state-of-the-art models for anomaly detection in video, including our teachers \citep{Georgescu-CVPR-2021, Georgescu-TPAMI-2021}, introduce temporal information as input to gain performance. Inspired by such methods, we explore the impact of using a temporal sequence of frames as input and present the results in Table~\ref{tab_frames}. We observe that the performance increases by 2-4\% for both AUC measures when using 3 video frames as input instead of a single frame. Going up to 5 frames as input does not seem to be further useful. To this end, we opt for providing 3 frames as input for the subsequent experiments.

\begin{table}[t]
\centering 
\setlength\tabcolsep{4.5pt}
\small{
\begin{tabular}{|c | c | c | c | c | c |} 
\hline
\multirow{2}{*}{Teachers} &   \multirow{2}{*}{$\mathcal{L}_{\mbox{\scriptsize{AE}}}$} & \multirow{2}{*}{$\mathcal{L}_{\mbox{\scriptsize{KD}}}$} & \multirow{2}{*}{$\mathcal{L}_{\mbox{\scriptsize{AKD}}}$} &  \multicolumn{2}{c|}{AUC}\\
 \cline{5-6}
 & & & & Micro & Macro  \\
 \hline
  \hline
$T_1$ &  \checkmark & & & 72.6 & 73.2 \\
$T_1$ & & \checkmark & &79.5 & 80.7 \\
$T_1$&   &  &\checkmark & 63.7 & 69.8 \\
$T_1$ & \checkmark & \checkmark & & 85.0 & 86.5 \\
$T_1$ &   \checkmark &  &\checkmark & 69.1 & 80.6 \\
$T_1$ &  & \checkmark  &\checkmark & 82.4  & 82.9 \\
$T_1$ & \checkmark & \checkmark & \checkmark & {86.2} & {86.7} \\
 \hline
$T_1$ $+$ $T_2$ & \checkmark & \checkmark & \checkmark & {88.3} & {87.9} \\
 \hline
\end{tabular}
}
\caption{Micro and macro AUC scores (in \%) on Avenue \citep{Lu-ICCV-2013}, while using different training procedures based on input reconstruction pre-training ($\mathcal{L}_{\mbox{\scriptsize{AE}}}$), knowledge distillation ($\mathcal{L}_{\mbox{\scriptsize{KD}}}$), and adversarial knowledge distillation ($\mathcal{L}_{\mbox{\scriptsize{AKD}}}$). Results with one versus two teachers are also included.}
\label{tab_losses} 
\end{table}

\begin{table}[t]
\centering 
\setlength\tabcolsep{3.2pt}
\small{
\begin{tabular}{| c | c | c | c | c | c | c | c |} 
\hline
   \multirow{4}{*}{\vspace{-0.18cm}Teachers} & \multirow{4}{*}{\vspace{-0.15cm}$\mathcal{L}_{\mbox{\scriptsize{AKD}}}$} & \multicolumn{6}{ c|}{AUC}  \\
   \cline{3-8}
   & &  \multicolumn{2}{c|}{Avenue} &  \multicolumn{2}{c|}{Shanghai} &  \multicolumn{2}{c|}{Ped2} \\
   \cline{3-8}
   &  & \rotatebox[origin=c]{90}{$\;$Micro$\;$} & \rotatebox[origin=c]{90}{Macro} & \rotatebox[origin=c]{90}{Micro} & \rotatebox[origin=c]{90}{Macro} & \rotatebox[origin=c]{90}{Micro} & \rotatebox[origin=c]{90}{Macro}\\
   \hline
    \hline
   $T_1$ 
   $+$ $T_2$ 
   & & {88.2} & {87.7} & 77.8 & 84.6 & 93.8 & 96.5\\
   $T_1$ 
   $+$ $T_2$ 
   & \checkmark & \textbf{88.3} & \textbf{87.9} & \textbf{78.0} & \textbf{86.0} & \textbf{95.9} & \textbf{97.1}\\
 \hline
\end{tabular}
}
\caption{Micro and macro AUC scores (in \%) on Avenue \citep{Lu-ICCV-2013}, ShanghaiTech \citep{Luo-ICCV-2017} and UCSD Ped2 \citep{Mahadevan-CVPR-2010}, for student models trained with two teachers, before and after adding adversarial distillation. Top scores are highlighted in bold.}
\label{tab_teachers_adv} 
\end{table}

\noindent{\bf Training objectives.}
We next ablate the proposed training procedure, which comprises two phases and three losses. In the first phase, we train the model to reconstruct the middle input frame, optimizing it with the loss $\mathcal{L}_{\mbox{\scriptsize{AE}}}$ defined in Eq.~\eqref{eq_ae}. In the second training phase, the model learns to distill knowledge from one or more teachers, optimizing the joint loss given by the sum of $\mathcal{L}_{\mbox{\scriptsize{KD}}}$ and $\mathcal{L}_{\mbox{\scriptsize{AKD}}}$, as defined in Eq.~\eqref{eq_total}. We study the effect of these three losses on Avenue using a single teacher, namely $T_1$ \citep{Georgescu-TPAMI-2021}, reporting the corresponding results in Table \ref{tab_losses}. First, we observe that training our model to reconstruct the middle frame is not sufficient to obtain competitive results on Avenue. Standard knowledge distillation obtains significantly better results than middle frame prediction and adversarial distillation. Combining the losses two by two brings improvements over each and every individual counterpart. Nevertheless, the results show that combining all three losses represents the training procedure achieving the best performance. This confirms that our training strategy is effective. In Table \ref{tab_losses}, we also present comparative results with one versus two teachers. The empirical comparison shows that using two teachers is better than using only one. To obtain a reliable assessment of the effect of adding adversarial distillation when using multiple teachers, we extend our ablation study to the ShanghaiTech \citep{Luo-ICCV-2017} and UCSD Ped2 \citep{Mahadevan-CVPR-2010} data sets. We report the corresponding results in Table \ref{tab_teachers_adv}. The empirical results on the additional data sets confirm our initial assessment observed on Avenue.  Although the impact of AKD is not major on Avenue, the performance gains are higher on ShanghaiTech and UCSD Ped2. Still, we observe that AKD consistently improves our student model on all datasets. Judging by the extended empirical evaluation on all three data sets, we conclude that adversarial knowledge distillation is generally useful. However, since the standard distillation process already performs very well, there is less room for improvement when introducing AKD. This explains why the potential gains generated by AKD are capped.


\begin{figure}[!t]
\begin{center}
\centerline{\includegraphics[width=0.68\linewidth]{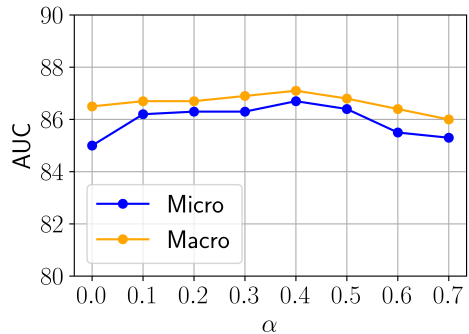}}
\vspace{-0.25cm}
\caption{Micro and macro AUC scores (in \%) on the Avenue data set \citep{Lu-ICCV-2013} for distinct values of the adversarial knowledge distillation weight factor $\alpha$.}
\label{fig_alpha}
\end{center}
\end{figure}

\noindent{\bf Adversarial loss weight.} 
Throughout all the experiments involving adversarial knowledge distillation, the hyperparameter controlling the importance of the adversarial distillation loss in Eq.~\eqref{eq_total} is set to $\alpha=0.1$. However, to determine if the proposed loss is robust to variations of $\alpha$, in Figure \ref{fig_alpha}, we present additional results with values of $\alpha$ between $0.1$ and $0.7$, taken at a step of $0.1$. The results for $\alpha=0$ correspond to the model trained without adversarial knowledge distillation. As indicated by the results illustrated in Figure \ref{fig_alpha}, there are multiple values for $\alpha$, between $0.1$ and $0.5$, leading to superior results compared with the baseline based only on standard knowledge distillation (which uses $\alpha=0$). We thus conclude that the loss proposed in Eq.~\eqref{eq_total} is robust to variations of $\alpha$, allowing us to set this hyperparameter without tuning.

\begin{table}[t]
\centering 
\setlength\tabcolsep{3.2pt}
\small{
\begin{tabular}{| c | c | c | c | c |} 
\hline
   \multirow{2}{*}{Teacher} & \#Adv.
   & \#Output & \multicolumn{2}{c|}{AUC}\\
   \cline{4-5}
    & Disc.
    & Branches & Micro & Macro\\
   \hline
    \hline
   $T_1$  & 0 & 1 & 86.2 & 86.7 \\
   $T_2$  & 0 & 1 & 82.1 & 85.5 \\
   $T_1$ $+$ $T_2$ & 0 & 1 & {88.2} & {87.7}\\
 \hline
   $T_1$ $+$ $T_2$ & 1 & 1 & 85.9 & 86.2 \\
   $T_1$ $+$ $T_2$ & 2 & 1 & \textbf{88.3} & \textbf{87.9} \\
   $T_1$ $+$ $T_2$ & 2 & 2 & 86.4 & 86.0 \\
   \hline
\end{tabular}
}
\caption{Micro and macro AUC scores (in \%) on the Avenue data set \citep{Lu-ICCV-2013}, while alternating between individual and combined teachers. Results for different attempts of performing adversarial distillation with multiple teachers are also reported. The top scores are highlighted in bold.}
\label{tab_teachers} 
\end{table}

We hereby underline that video anomaly detection data sets do not have separate validation sets (keeping a part of the training set is also not possible due the lack of anomalies at training time), and it would be unfair to tune $\alpha$ on the test set. We thus set $\alpha$ to $0.1$ without tuning. For the same reason, we refrain from adding more hyperparameters, \eg~we simply sum the losses with respect to all teachers in Eq.~\eqref{final_pixel} instead of considering a weighted sum.

\noindent{\bf Multiple teachers and adversarial training.}
The ablation results presented so far distill knowledge from a single teacher, namely $T_1$ \citep{Georgescu-TPAMI-2021}. Since different models can often produce distinct anomaly maps, it can be reasonably argued that combining the respective models can lead to performance improvements, an argument which forms the basis of ensemble learning \citep{Rokach-AIR-2010}. We conjecture that the same principle applies to knowledge distillation, \ie~distilling knowledge from multiple teachers is likely to bring performance gains. To this end, we propose to distill knowledge from two powerful teachers, denoted as $T_1$ \citep{Georgescu-TPAMI-2021} and $T_2$ \citep{Georgescu-CVPR-2021}. In Table \ref{tab_teachers}, we present results for alternating between individual and combined teachers, applying standard knowledge distillation at first. Our empirical results confirm that introducing two teachers leads to superior micro and macro AUC scores. 

We next turn our attention to introducing adversarial knowledge distillation with multiple teachers, which is not trivial. We study three possible ways of integrating multiple teachers with adversarial distillation:
\begin{itemize}
    \item \vspace{-0.0cm}{\bf One adversarial discriminator, one output branch.} In this configuration, the student has a single multi-resolution output branch for both teachers, and the adversarial discriminator distinguishes between three categories of anomaly maps produced by the student, the teacher $T_1$, and the teacher $T_2$, respectively. Within this framework, adding more teachers is just a matter of adding more classes to the discriminator. 
    \item \vspace{-0.0cm}{\bf Two adversarial discriminators, one output branch.} In this configuration, the student has a single multi-resolution output branch for both teachers. For each teacher $T_i$, there is a distinct binary discriminator that classifies the anomaly maps into two classes, one for the student and one for the teacher $T_i$. Within this framework, adding more teachers requires adding an equal number of adversarial discriminators.
    \item \vspace{-0.0cm}{\bf Two adversarial discriminators, two output branches.} In this framework, the student has a separate multi-resolution output branch and an associated adversarial discriminator for each teacher $T_i$. Since there are multiple output branches for this framework, we need a method to aggregate the anomaly maps at a certain resolution. We consider taking the element-wise maximum or the average of the corresponding anomaly maps, reporting the results with the better aggregation function only, namely the element-wise maximum. 
\end{itemize}

In the second part of Table \ref{tab_teachers}, we present the results with the aforementioned frameworks for integrating multiple teachers with adversarial distillation. Among the three alternative frameworks, the best results are obtained by the approach comprising two adversarial discriminators and one output branch. However, it appears that the other two frameworks are not able to surpass the results obtained by distilling both teachers with standard knowledge distillation. In summary, there seems to be only one way to go forward: using two adversarial discriminators and one output branch. Still, the performance gains of the only viable approach on Avenue seem marginal. In Table \ref{tab_teachers_adv}, we have already extended this analysis to ShanghaiTech and UCSD Ped2, observing that combining adversarial distillation and multiple teachers brings higher gains on the other benchmarks. All in all, these results confirm that it requires meticulous design and significant effort in formulating a unified framework that is \textit{both} efficient and effective.

\begin{table}[t]
\centering 
\setlength\tabcolsep{3.2pt}
\small{
\begin{tabular}{| c | c | c |} 
\hline
   \multirow{2}{*}{Teacher} & \multicolumn{2}{c|}{AUC}\\
   \cline{2-3}
    & Micro & Macro\\
   \hline
    \hline
   $T_1$ & 86.2 & 86.7 \\
   $T_2$ & 82.1 & 85.5 \\
 \hline
    $T_1$ $+$ $T_1$  & {86.7} & {87.1} \\
   $T_1$ $+$ $T_2$ & {88.3} & {87.9} \\
   
   $T_1$ $+$ $T_2$ $+$ $T_3$ & \textbf{88.5} & \textbf{88.2} \\
   \hline
\end{tabular}
}
\caption{Micro and macro AUC scores (in \%) on the Avenue data set \citep{Lu-ICCV-2013}, while testing various teacher combinations. The top scores are highlighted in bold.}
\label{tab_more_teachers} 
\end{table}

\begin{table}[t]
\centering 
\setlength\tabcolsep{3.2pt}
\small{
\begin{tabular}{| l | c | c | c |} 
\hline
  \multirow{2}{*}{Architecture} & \multicolumn{2}{c|}{AUC}  & \multirow{2}{*}{FPS}\\
   \cline{2-3}
    & Micro & Macro & \\
     \hline
      \hline
  CNN $+$ CvT (standard, with fc) & 71.8 & 77.1 & 1267 \\ 
  CNN $+$ CvT (pointwise conv, no fc) & \textbf{85.0} & \textbf{86.5} & \textbf{1480}\\
 \hline
\end{tabular}
}
\caption{Micro and macro AUC scores (in \%) and FPS rates on the Avenue data set \citep{Lu-ICCV-2013}, while alternating between CvT blocks with fully connected (fc) layers (standard) or pointwise convolutions (proposed). The top scores are highlighted in bold.}
\label{tab_architecture} 
\end{table}

\noindent{\bf Teacher variations.} 
Upon establishing how to integrate adversarial distillation with multiple teachers, we test additional hypotheses regarding the number and the diversity of teachers. We report the corresponding results in Table \ref{tab_more_teachers}. First, we carry out an experiment with two runs of teacher $T_1$ on Avenue, aiming to verify if the diversity among teachers is important. We obtain inferior performance when distilling two runs of teacher $T_1$ instead of teachers $T_1$ and $T_2$. This indicates that the diversity of teachers is an important factor influencing the robustness of our student. We thus conclude that the effectiveness of the ensemble is more correlated with the diversity and complementary nature of the teachers, rather than the sheer number of teachers. This is because employing distinct checkpoints of the same teacher brought a limited performance boost, while employing more diverse teachers ($T_1$ and $T_2$) raises the performance by a significant margin.

Next, we conduct another experiment on Avenue by adding a new teacher $T_3$ \citep{Liu-ICCV-2021} to our lineup of teachers. The teacher developed by \citet{Liu-ICCV-2021} stimulates our student to reach even better results. However, the performance gains are not as high as going from one teacher to two teachers. Therefore, it is likely to observe a saturation of the improvements after a certain number of teachers.

\noindent{\bf Dense layers versus pointwise convolutions.}
In Table~\ref{tab_architecture}, we demonstrate the effectiveness of replacing the last two fully connected layers from the CvT block \citep{Wu-ICCV-2021} with pointwise convolutions. For a fair comparison, both model variants are trained with middle frame reconstruction in the first training stage and standard knowledge distillation in the second stage. Aside from the expected speed improvement, we observe that the proposed change also generates significantly better micro and macro AUC scores. In conclusion, the empirical evaluation clearly indicates that replacing fully connected layers with pointwise convolutions is beneficial in terms of both speed and accuracy.

While replacing the multi-layer perceptron with pointwise convolutions might seem as a natural choice, to our knowledge, this has not been proposed before within anomaly detection transformers. We consider this contribution very important due to its accuracy and speed gains ($+13.2\%$ micro AUC and $+213$ FPS).

\section{Reproducibility}

To ensure reproducibility, our source code is publicly released at \url{https://github.com/ristea/fast-aed}. With the same goal in mind, we next present the student architecture in more detail.

Our student comprises a convolutional downsampling block, a sequence of transformer blocks, and a set of output blocks. The downsampling block has five convolutional layers. The first convolutional layer is formed of $16$ filters with a spatial size of $7\times 7$ and no padding. The subsequent convolutional layers of the downsampling block have $32$, $64$, $128$ and $256$ filters, respectively, each with a kernel size of $3\times 3$. Each layer uses a stride of $2$ and a padding of $1$.

The transformer encloses $m$ consecutive transformer blocks and interprets $P$ as a set of overlapping visual tokens. The transformer blocks follow the implementation of \citet{Wu-ICCV-2021}, where, prior to the self-attention mechanism, the queries $Q$, the keys $K$ and the values $V$ are computed from $P$ via convolutional projection. This operation is a depthwise separable convolution, being implemented with different weights for $Q$, $K$ and $V$. The depthwise convolution has $256$ filters with a spatial dimension of $3\times 3$. It is applied at a stride of $2$ for $K$ and $V$, and a stride of $1$ for $Q$. This layer is followed in all three cases by a batch normalization layer and a pointwise convolution with $64$ filters. The final matrices $Q \in \mathbb{R}^{n_q \times d}$, $K \in \mathbb{R}^{n_k \times d}$ and $V \in \mathbb{R}^{n_v \times d}$ are obtained after flattening the activation maps, while preserving the channel dimension. We note that $n_q = h \cdot w$, $n_k = n_v = \frac{h \cdot w}{4}$ and $d=64$.

The next stage in the transformer block is the self-attention mechanism defined as follows:
\begin{equation}
    \begin{split}
        Z = \mbox{\emph{softmax}}\left(\frac{
        Q\cdot K^\top}{\sqrt{d_q}}\right)\cdot V,
    \end{split}
\end{equation}
where $K^\top$ is the transpose of $K$, and $Z$ is a matrix of size $n_q\times d$ containing the new value vectors. We reshape $Z$ into a tensor of $h\times w\times d$ components.

Our network comprises a multi-head attention module, each head containing one convolutional projection and a self-attention mechanism. The tensors resulting from all heads are concatenated along the channel axis, obtaining a tensor $Z^{\circ}$ of $h\times w \times (d\cdot s)$ components, where $s$ is the number of heads. The tensor $Z^{\circ}$ is further passed through a pointwise convolutional layer that reduces the number of activation maps to the number of channels $c$. The tensor $P$ goes through a skip connection, being summed up with $Z^{\circ}$. The resulting output is denoted by $Z^*$. 
The final output of the multi-head attention module is fed into a batch normalization layer. To significantly improve efficiency, we replace the multi-layer perceptron that typically follows in a conventional transformer block with two pointwise convolutional layers. The first pointwise convolutional layer comprises $4\cdot c$ filters, and the second one only $c$ filters. Another skip connection is added between the input $Z^*$ and the output of this module.

\section{Limitations}

Although we achieved a very good trade-off between efficiency and performance in this work, we admit that there is an observable gap between the performance of our model and that of the best-performing object-centric method. Moreover, since our method is based on knowledge distillation, we also need to apply the teachers and obtain target anomaly maps, whenever we want to train our method on a new dataset. 
In future work, our objective is to enhance performance, which can be pursued by introducing more proxy tasks for the student, in compliance with those used by the teachers. Moreover, investigating the potential of pre-training approaches via self-supervised tasks, in conjunction with various architectural configurations, represents an additional compelling direction of future research.

\section{Conclusion}

In this paper, we introduced a novel teacher-student framework for anomaly detection in video, learning to detect anomalies by distilling knowledge from multiple highly accurate object-level teachers into a highly efficient student. Moreover, we proposed adversarial knowledge distillation in the
context of anomaly detection, introducing an adversarial discriminator for each teacher. Using our novel framework, we successfully trained an efficient convolutional transformer, obtaining a state-of-the-art speed versus performance trade-off. More precisely, we reached an unprecedented speed of 1480 FPS, with a minimal performance decrease. 

In future work, we aim to further raise the accuracy of our student by introducing more teachers and tasks.

\section*{Acknowledgements}
This work was supported by a grant of the Romanian Ministry of Education and Research, CNCS - UEFISCDI, project number PN-III-P2-2.1-PED-2021-0195, within PNCDI III.



\bibliographystyle{model2-names}
\bibliography{refs}


\end{document}